\documentclass[runningheads]{llncs}

\usepackage{eccv}

\usepackage{eccvabbrv}

\usepackage{graphicx}
\usepackage{booktabs}
\usepackage{tcolorbox}
\tcbuselibrary{skins,breakable}

\usepackage[accsupp]{axessibility}  
\usepackage[table]{xcolor}

\definecolor{bestgreen}{RGB}{180,235,180}
\definecolor{secondgreen}{RGB}{220,245,220}

\usepackage[pagebackref,breaklinks,colorlinks,citecolor=eccvblue]{hyperref}

\usepackage{orcidlink}
\newtcolorbox{questionbox}{
  enhanced,
  breakable,
  colback=white,
  colframe=black,
  boxrule=0.8pt,
  arc=6pt,
  left=10pt,
  right=10pt,
  top=8pt,
  bottom=8pt
}

\usepackage{booktabs}
\usepackage{multirow}
\usepackage{graphicx}
\usepackage{pifont} 
\usepackage{array}

\newcommand{\cmark}{\ding{51}}
\newcommand{\xmark}{\ding{55}}

\begin{document}

\title{CVSBench: A Comprehensive Benchmark for Cross-view Spatial Reasoning and Dreaming } 

\titlerunning{CVSBench}


\author{Ruixun Liu\inst{1*} \and Lingyu Zhang\inst{1*} \and  Lanxuan Xue\inst{1*}\and Kaiyu Li\inst{1,\dag} \and \\Bowen Fu\inst{1} \and Xiangyong Cao\inst{1,2,\dag}
}

\authorrunning{Ruixun Liu et al.}

\institute{\textsuperscript{1}Faculty of Electronic and Information Engineering, Xi’an Jiaotong University\\
\textsuperscript{2}Ministry of Education Key Laboratory of Intelligent Networks and Network Security\\}

\maketitle

\renewcommand{\thefootnote}{*}
\footnotetext[1]{Equal contribution.}
\renewcommand{\thefootnote}{\dag}
\footnotetext[1]{Corresponding author (likyoo.ai@gmail.com, caoxiangyong@mail.xjtu.edu.cn)}
\renewcommand{\thefootnote}{\arabic{footnote}}

\begin{abstract}
 Humans can effortlessly reason about scenes across different viewpoints, yet it remains unclear whether Vision–Language Models (VLMs) possess similar cross-view spatial abilities. Satellite-street scene pairs, with their complex contexts and extreme viewpoint variations, provide an ideal testbed. Motivated by this, we introduce CVSBench, a large-scale benchmark for evaluating cross-view spatial reasoning through satellite-street pairs.
 This benchmark supports multiple tasks, including cross-view VQA, cross-view grounding, and viewpoint identification. CVSBench comprises 3,297 cross-view image groups with 9,468 object-level annotations and 40,679 question–answer (QA) pairs, enabling systematic and controlled evaluation of cross-view spatial reasoning. Extensive evaluations reveal that advanced VLMs struggle to maintain object-level and layout consistency under drastic viewpoint changes. To bridge this gap towards human-like spatial cognition, we investigate two categories of approaches: spatially grounded reasoning and the incorporation of cognitive map inputs.
 Our findings demonstrate that language-only reasoning yields marginal improvements, while incorporating visual spatial imagination via a 3D scene imagination pipeline substantially improves cross-view reasoning. These results highlight the necessity of explicit visual-spatial representations for robust spatial cognition in VLMs. Our data and code are released at \url{https://huggingface.co/datasets/zlyzlyzly/CVSBench}.
\end{abstract}

\begin{figure}[t]
    \centering
    \includegraphics[width=1.0\linewidth]{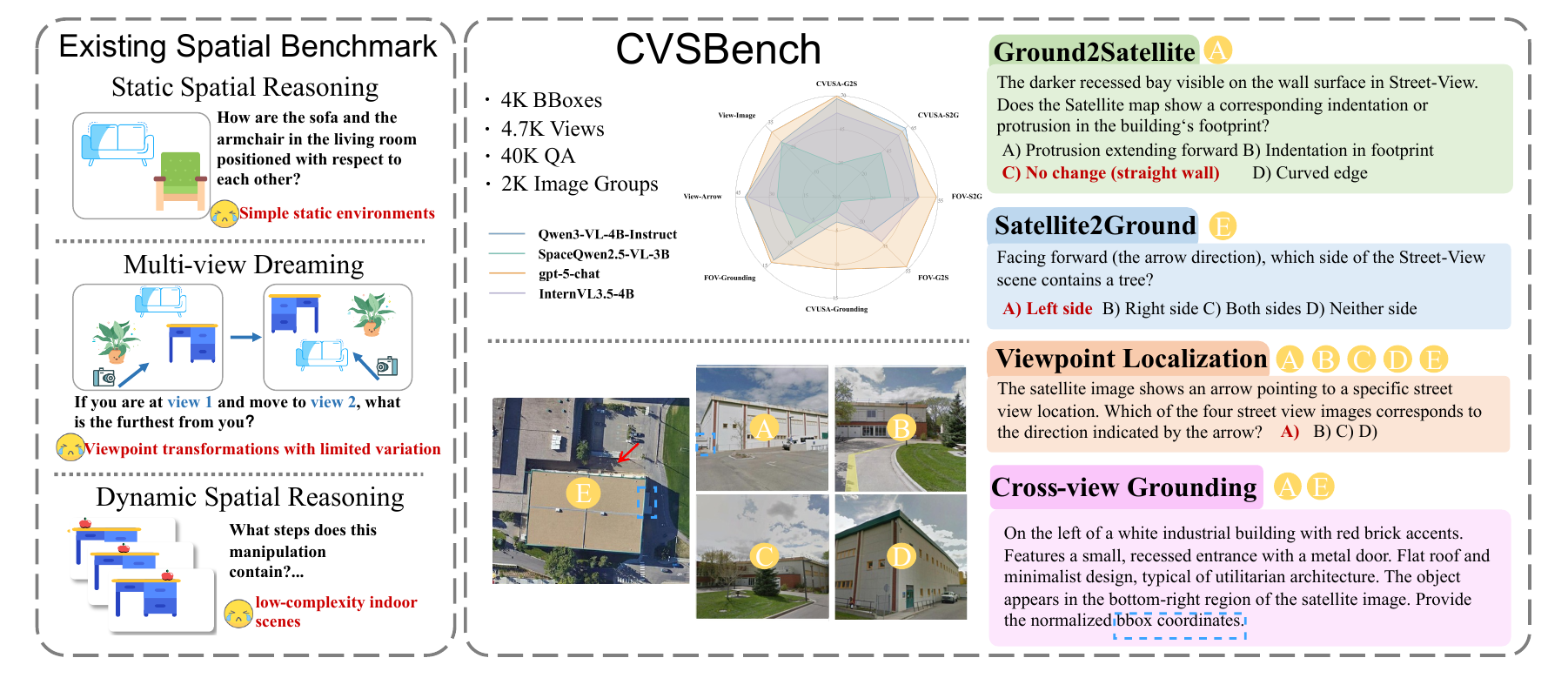}
    \vspace{-6mm}
    \caption{Overview of \textbf{CVSBench}. The left panel reviews representative tasks from existing spatial benchmarks suffering from several limitations~\cite{szymanska2024space3d, jia2025omnispatial, yin2025spatial}. The right panel displays our benchmark specifications, a radar chart comparing model performance, and core tasks with sample questions, demonstrating the superiority of our benchmark in task diversity, scale, and difficulty. The indices (e.g., A–E) shown after each subtask title indicate the input images for that task. A red arrow indicates the viewing perspective, and blue BBoxes denote the target object.}
    \label{intro}
    \vspace{-5mm}
\end{figure}

\section{Introduction}
\label{sec:intro}

Vision-Language Models (VLMs)~\cite{hurst2024gpt, bai2025qwen3, bai2025qwen25vltechnicalreport, wang2025internvl3, comanici2025gemini} have been extensively studied for spatial reasoning recently~\cite{chen2024spatialvlm, liu2023vsr, yu2025far, cheng2024spatialrgpt}. However, with the rapid development of embodied navigation, autonomous driving, and urban scene perception, static spatial understanding is no longer sufficient to meet the demands of these tasks~\cite{ding2024holistic, zhu2025move, ji2025robobrain}. Humans, for instance, can effortlessly imagine what a street-level scene would look like from an overhead view, or integrate multiple egocentric observations of object distributions into a coherent map-centric layout~\cite{newcombe2014thinking, bednarz2019improves}. Even across different coordinate systems or viewing perspectives, humans are able to maintain a consistent internal representation of spatial relationships. Despite the impressive progress achieved by modern VLMs, they still struggle to imagine a scene from an alternative viewpoint given a single observation, as well as to perform reliable cross-view matching of the same objects~\cite{yin2025spatial, gholami2025spatial}.

To overcome this limitation, recent studies begin to explore the capability of VLMs in understanding dynamic or multi-view scenes~\cite{lee2025perspective, yang2025thinking, yin2025spatial, ji2025robobrain, li2025viewspatial}. Nevertheless, these efforts still suffer from two major limitations: (1) existing benchmarks are primarily constructed from indoor objects or simple scenes, limiting cross-view evaluation in complex real-world environments~\cite{ji2025robobrain, yu2025far, azuma2022scanqa}; and (2) multi-view settings are often restricted to cameras rotating around an object or the agent itself, resulting in relatively small viewpoint variations~\cite{yin2025spatial, yang2025thinking}. 
To address these challenges, we adopt satellite–street view scenarios~\cite{toker2021coming, wang2025geovista, zhu2021vigor}, which naturally exhibit complex urban layouts and large viewpoint discrepancies between ground-level and overhead observations.
As illustrated in Fig.~\ref{intro}, reasoning in satellite–street view scenarios requires a set of core spatial capabilities, including
(i) imagining observations from unseen viewpoints given a single-view input (\textit{e.g.}, whether concave structures on building sides are visible from an overhead view); 
(ii) maintaining cross-view consistency (\textit{e.g.}, aligning the same objects across views); and 
(iii) inferring camera viewpoints based on visual cues.

Based on these requirements, we introduce CVSBench, a large-scale benchmark for cross-view spatial reasoning and imagination, comprising the FOV-subset and CVUSA-subset. 
To obtain accurate annotations and comprehensive evaluation tasks, we design a semi-automatic annotation pipeline that labels cross-view object Bounding Boxes (BBoxes) as well as multi-view alignment relationships. Moreover, we exploit differences in the observation difficulty of the same object across viewpoints to construct QA pairs that probe the spatial imagination of models under limited views. Using this pipeline, CVSBench comprises 3,297 image groups and supports a diverse set of evaluation tasks, including cross-view visual question answering (VQA), grounding, and viewpoint identification. CVSBench enables a systematic evaluation of the spatial cognition in general-purpose VLMs under cross-view scenarios.

Using CVSBench, we conduct extensive evaluations on a wide range of state-of-the-art VLMs and find that existing models still exhibit notable limitations in cross-view scenarios. Inspired by prior work on VLM reasoning and reasoning with visual representations~\cite{zhang2024if, yang2025machine, wu2024mind, lee2025perspective}, we raise the following question:

\begin{questionbox}
\textbf{Question.}
\textit{
Can VLMs be endowed with stronger cross-view spatial understanding abilities by incorporating explicit reasoning strategies or auxiliary perceptual processes?
}
\end{questionbox}

To address this question, we first investigate whether training-free chain-of-thought (CoT) prompting can lead to performance improvements. Building on this analysis, we further propose two CoT-based strategies.
(1) Structured Scene CoT, which parses the visual input into a structured textual format. This strategy requires the model to explicitly enumerate object categories and their spatial distributions within the current view before answering the question.
(2) Spatial Imagination CoT, which simulates the perspective transformation. This strategy encourages the model to describe the scene composition and layout from the unavailable target viewpoint through textual inference. After training with Supervised Fine-Tuning (SFT) and Reinforcement Learning (RL), the overall gains remain modest, with an average improvement of approximately 1.75\% across categories on the FOV-subset, while the improvement on CVUSA-subset is only around 0.4\%. Current models struggle to reason about and imagine spatial layouts from alternative viewpoints using textual reasoning alone.

Furthermore, we introduce a 3D scene imagination strategy. Since most existing general-purpose VLMs lack the capability to generate visual intermediate representations~\cite{chen2024spatialvlm, cheng2024spatialrgpt}, we employ depth estimation models and image generation models to synthesize images from 3D viewpoints. These generated images simulate the depth-centered features and ``God’s-eye-view'' cognitive map representations that humans construct mentally. Empirical evaluations demonstrate that, on the FOV-subset, incorporating depth-estimated images leads to a 1.23\% improvement, while cognitive map representations yield a 3.34\% gain.
Our empirical evidence reveals an important finding: compared with purely text-based reasoning approaches, visual imagination in VLMs has the potential to bring substantially greater improvements to cross-view spatial reasoning tasks.

Our main contributions are summarized as follows:
\begin{itemize}
    \item We are the first to integrate the \textbf{satellite - street view} cross-view VQA task with geo-localization task, enabling a more comprehensive study of the cross-view spatial reasoning and imagination capabilities of VLMs.
    \item We propose \textbf{CVSBench}, a large-scale benchmark dataset featuring extensive human annotations and comprehensive QA pairs, enabling systematic evaluation across multiple cross-view spatial reasoning tasks.
    \item Through a series of experiments, we demonstrate that the cross-view spatial understanding of current models is primarily constrained by their inability to reason about spatial distributions under viewpoint transformation, and that it can be effectively enhanced through visual spatial imagination strategies.
\end{itemize}

\section{Related Work}
\label{sec:related}
\subsection{Spatial Thinking evaluation benchmarks}

The evaluation for spatial intelligence is undergoing a profound transformation from simple VQA to systematic cognitive assessment, an evolution deeply inspired by theories of spatial thinking components in cognitive psychology~\cite{lee2012components, johnson1980mental, bednarz2019improves, newcombe2014thinking}. Early benchmarks are primarily confined to 2D relationship judgments~\cite{liu2023vsr, fu2024blink} or single-modal geometric attribute perception~\cite{azuma2022scanqa, ma20253dsrbench, cheng2024spatialrgpt}. To address this limitation, the new generation of benchmarks emphasizes cognitive hierarchy and dynamic interaction: recently proposed SIBench~\cite{yu2025far} and OmniSpatial~\cite{jia2025omnispatial} introduce comprehensive classifications covering perception, understanding, and planning; meanwhile, SPACE~\cite{ramakrishnan2024does}, systematically examines spatial abilities ranging from navigation to object manipulation. Moreover, VSI-Bench~\cite{yang2025thinking} evaluates spatiotemporal memory via video streams, while MINDCUBE~\cite{yin2025spatial} further focuses on spatial consistency and mental model construction under multi-view settings. However, existing benchmarks are largely limited to indoor environments, lacking urban-scale evaluations for cross-view localization and logical reasoning between overhead and ground-level perspectives.

\subsection{Spatial Understanding and Reasoning in VLMs}

While general VLMs have demonstrated remarkable proficiency in semantic understanding and open-ended generation~\cite{alayrac2022flamingo, li2023blip, liu2023visual, dai2023instructblip, liu2024improved, hurst2024gpt, wang2024qwen2, bai2025qwen3}, they still suffer from intrinsic deficiencies in precise spatial perception due to the lack of explicit grounding in the physical world, often exhibiting severe ``egocentric bias'' or ``spatial hallucinations''~\cite{chen2024spatialvlm, guan2024hallusionbench, li2023evaluating, tong2024eyes, fu2023mme}. To enhance spatial reasoning, some studies synthesize large-scale perception data~\cite{chen2024spatialvlm} or apply Multimodal CoT mechanisms~\cite{zhang2023multimodal, mitra2024compositional}. Other works inject geometric priors by introducing 3D features~\cite{hong20233d, wang2023chat} or leveraging visual geometry foundation models~\cite{wu2025spatial}. Recently, human-like mental simulation has been explored for spatial tasks. For instance, embodied agents predict physical trajectories~\cite{ji2025robobrain}, and perspective-taking models use mental imagery for cross-view deduction~\cite{lee2025perspective}. Additionally, explicit cognitive maps are constructed to help reason under limited viewpoints~\cite{yin2025spatial}. However, they currently lack the capacity for ``urban-scale'' cognition~\cite{kim2024openvla, zhu2024llava, yuan2024robopoint}, particularly in processing cross-view data from satellite and street views, which restricts their applicability in broader out-door tasks.

\subsection{Remote Sensing VQA and Geo-Grounding tasks}

In recent years, multi-modal remote sensing research has achieved remarkable progress in VQA~\cite{lobry2020rsvqa, hu2025rsgpt} and deep semantic understanding of single-view or cross-view contexts~\cite{ muhtar2024lhrs, li2024vrsbench, xu2026citycube}. Building upon these advances, cross-view geo-localization has also evolved from basic feature matching toward complex spatial reasoning that integrates street-view and satellite images~\cite{zhan2023rsvg, wang2025geovista, zhu2021vigor, toker2021coming}. However, current remote sensing visual question answering and cross-view localization remain greatly isolated from each other.
 To bridge this gap, we present the integration of cross-view localization with VQA, incorporating spatial reasoning to provide a more comprehensive exploration of geospatial intelligence.

\begin{figure}[t]
    \centering
    \includegraphics[width=0.9\linewidth]{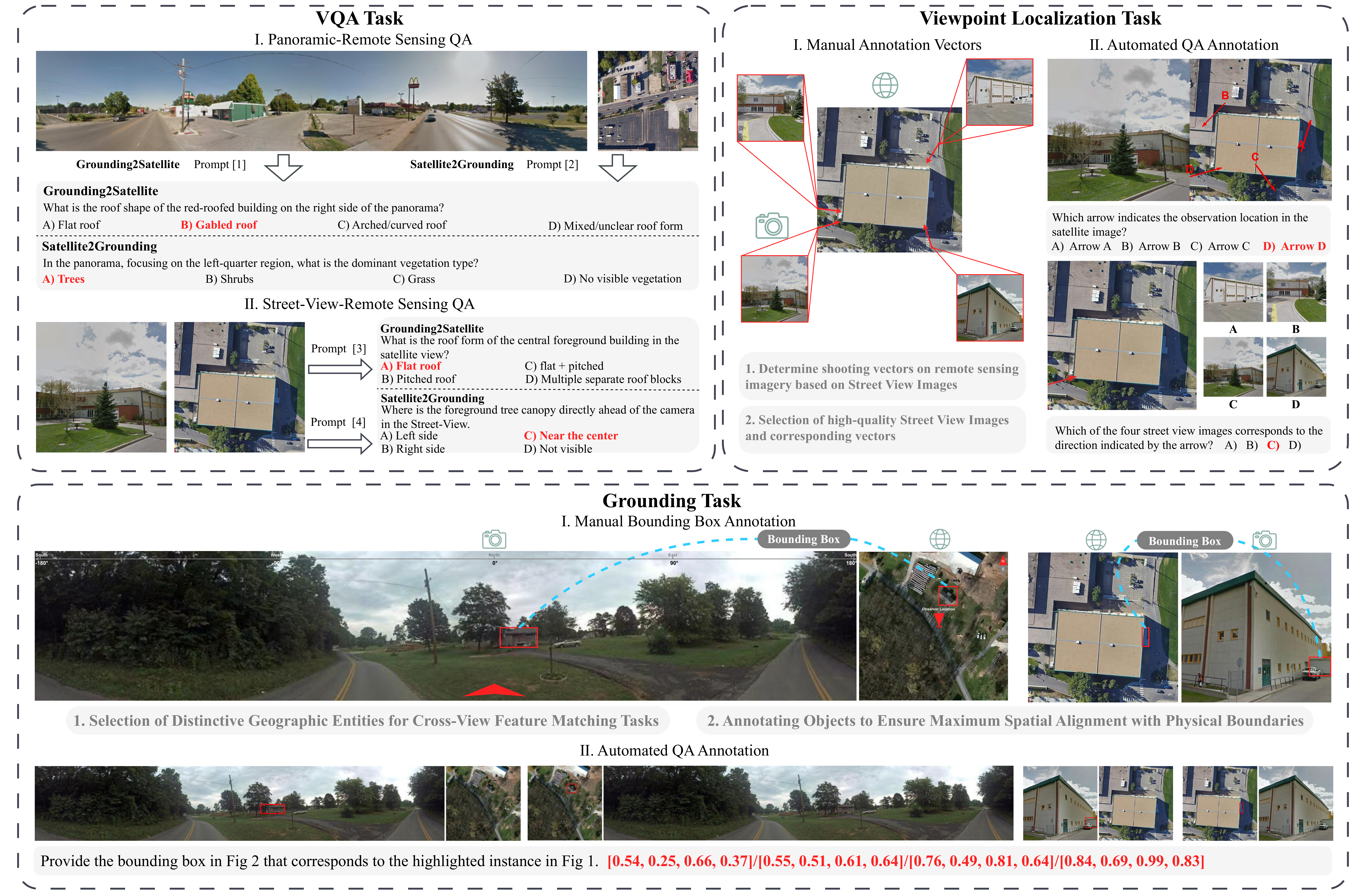}
    \vspace{-2mm}
    \caption{Dataset annotation pipeline (Red: ground truth (GT)). \textbf{VQA}: Multi-type prompts covering diverse question formats and instruction styles. \textbf{Viewpoint Localization}: Red arrows denote camera poses. \textbf{Grounding}: Red bboxes, arrows, and dots represent objects, viewpoints, and camera locations.}
    \label{pipeline}
    \vspace{-4mm}
\end{figure}

\section{Benchmark and Evaluation}
\label{Benchmark}
We introduce \textbf{CVSBench}, a benchmark for evaluating spatial reasoning  across heterogeneous viewpoints and comprises three task types: \emph{cross-view VQA}, \emph{cross-view grounding}, and \emph{viewpoint selection}.

\subsection{Data Annotation Pipeline}

CVSBench is constructed from two cross-view datasets: CVUSA~\cite{cvusa} and University1652~\cite{university1652}. We curate 2,155 satellite--panorama image pairs from CVUSA (CVUSA-subset) and extract 1,142 satellite-street view image pairs from University1652 (FOV-subset). A semi-automatic annotation pipeline as shown in Fig.~\ref{pipeline} organizes the data into three task families with critical human verification.

\paragraph{Cross-View VQA.}
The VQA task is structured into two inverse modes: Ground-to-Satellite (G2S) and Satellite-to-Ground (S2G). Structured prompt templates are constructed and combined with paired images to generate candidate questions using \textit{gemini-2.5-flash}~\cite{comanici2025gemini}, covering attributes such as visibility, connectivity, object properties, and spatial relations.
Importantly, the two viewpoints provide asymmetric spatial cues: certain attributes are directly observable and easily answered in one view, while they become implicit, partially occluded, or geometrically ambiguous in the other. We leverage this asymmetry to design both the annotation and evaluation protocol. During annotation, the model has access to image pairs to ensure semantic correctness and cross-view consistency. During evaluation, however, it is restricted to a single-view input, requiring the model to infer the missing spatial information through cross-view geometric reasoning rather than direct visual evidence.

\paragraph{Viewpoint Localization.}
This task is defined only for the FOV-subset. Human annotators mark 2D directional arrows on satellite images, where the start and end points represent camera location and viewing orientation. Based on these geometric priors, two tasks are constructed:
(1) View-Arrow, which requires selecting the correct camera pose arrow in a satellite-view image given a street-view image; and
(2) View-Image, which requires selecting the correct street-view image given a satellite image with a fixed arrow.

\begin{figure}[t]
    \centering
    \includegraphics[width=0.9\linewidth]{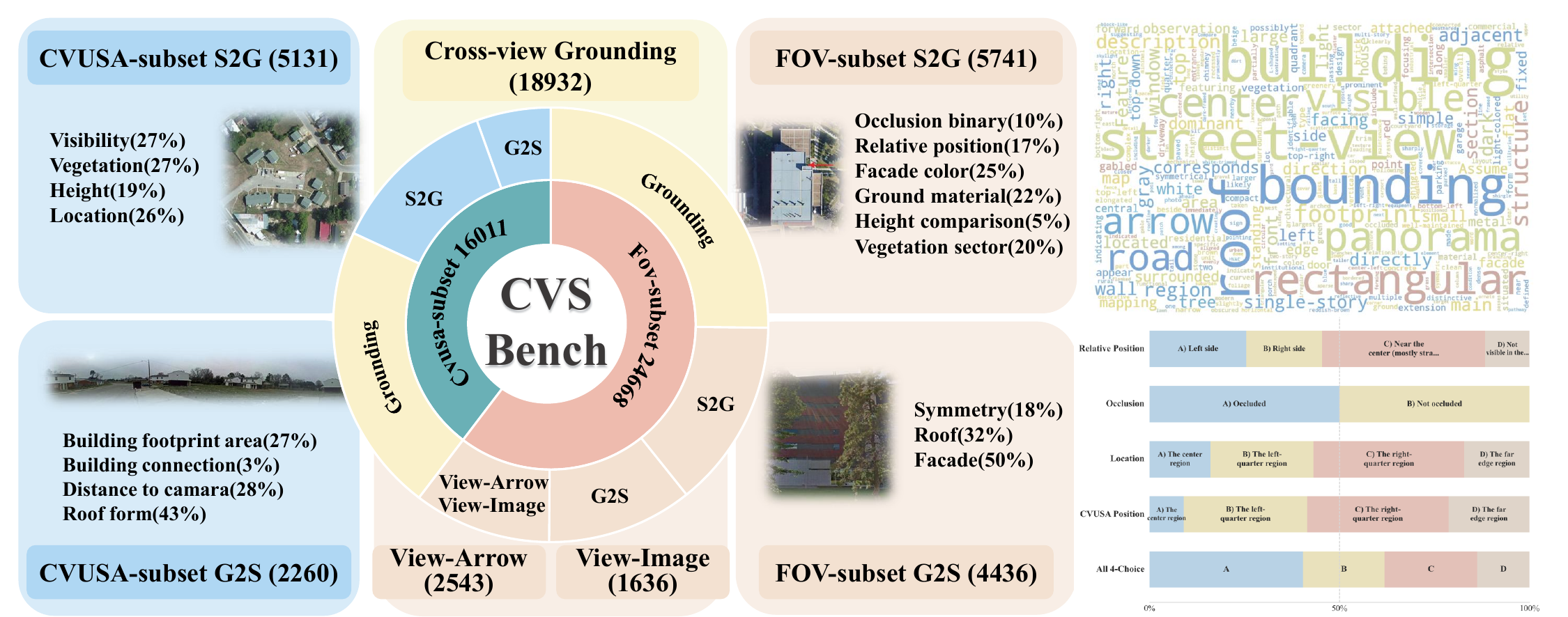} 
    \vspace{-2mm}
\caption{ 
Overview of task categories in CVSBench. The left panel shows the detailed categorization of task types in CVSBench, with counts or proportions indicated in parentheses. The right panel presents a word cloud of the QA composition and the distribution of typical answer categories.
}
\vspace{-4mm}
    \label{fig:cvsbench}
\end{figure}

\paragraph{Cross-View Grounding.}
For the CVUSA-subset, BBox of major objects are first generated in the satellite view using \textit{gemini-2.5-flash}~\cite{comanici2025gemini}. 
Based on the object location, a corresponding line-of-sight direction is estimated to search for candidate regions in the ground view, forming an initial cross-view correspondence which is then manually reviewed and corrected. 
For the FOV-subset, BBoxes are manually annotated. Evaluation includes two formulations:
(1) \emph{BBox-to-BBox}: given a BBox in view A, predict the corresponding BBox in view B;
(2) \emph{Description-guided grounding}: given a textual description from view A and a coarse region in view B, predict the BBox in view B.

Questions that can be answered without images are filtered out, after which the remaining instances undergo human verification. The primary review criteria include checking cross-view identifiability and consistency, as well as correcting incorrect answers. To mitigate language and answer biases introduced during the QA generation process, we analyze and balance the distribution of different option labels. Furthermore, since the evaluation is under single-view settings, the model is forced to reason based on information from a limited viewpoint, thereby reducing potential response bias. Finally, eight professional annotators spend approximately 100 hours on annotation and verification, and about 30\% of the QA pairs are manually corrected. More details are provided in the Appendix.

\begin{table*}[t]
\centering
\caption{Comparison of CVSBench with existing datasets. The domain categories include RS, General Domain (Gen), and General Spatial (Gen-S). Hyphens (-) denote missing values in the training set. \textbf{TI}: Text Instruction, \textbf{TA}: Text Answer, \textbf{VP}: Visual Prompt. For task capabilities, \cmark~indicates support, while \xmark~indicates lack thereof.}
\vspace{-2mm}
\label{tab:dataset_comparison}
\resizebox{\textwidth}{!}{%
\begin{tabular}{c l c c c c c c c c}
\toprule
\multirow{2}{*}{\textbf{Category}} & \multirow{2}{*}{\textbf{Benchmark}} & \multirow{2}{*}{\textbf{Train}} & \multirow{2}{*}{\textbf{Test}} & \multirow{2}{*}{\textbf{Anno. Format}} & \multirow{2}{*}{\textbf{Evaluated Capabilities}} & \multirow{2}{*}{\textbf{Query Format}} & \multicolumn{3}{c}{\textbf{Task Capabilities}} \\
\cmidrule(lr){8-10}
& & & & & & & \textbf{VQA} & \textbf{Grounding} & \textbf{Multi-view} \\
\midrule

\multirow{5}{*}{RS} 
& RSVQA~\cite{lobry2020rsvqa} & 715K & 428K & TI, TA & Basic perception \& counting & Short Answer & \cmark & \xmark & \xmark \\
& RSVGD~\cite{zhan2023rsvg} & 23K & 15.3K & TI, BBox & Attribute-based grounding & Phrases & \xmark & \cmark & \xmark \\
& RSIEVAL~\cite{hu2025rsgpt} & - & 936 & TI, TA & Scene captioning \& reasoning & Open-ended & \cmark & \xmark & \xmark \\
& LHRS-Bench~\cite{muhtar2024lhrs} & - & 690 & TI, TA & Multi-dim semantic understanding & MCQ & \cmark & \xmark & \xmark \\
& VRSBench~\cite{li2024vrsbench} & 85.8K & 37.4K & TI, TA, BBox & OBB grounding \& captioning & Open-ended & \cmark & \cmark & \xmark \\
\midrule

\multirow{5}{*}{Gen}
& MME~\cite{fu2023mme} & - & 2.8K & TI, TA & Perception \& cognition & T/F & \cmark & \xmark & \xmark \\
& MMBench~\cite{liu2024mmbench} & - & 3.2K & TI, TA & Fine-grained logical reasoning & MCQ & \cmark & \xmark & \xmark \\
& SEED-Bench~\cite{li2024seed} & - & 24K & TI, TA & Hierarchical comprehension & MCQ & \cmark & \xmark & \cmark \\
& VisIT-Bench~\cite{bitton2023visit} & - & 592 & TI, TA & Instruction following & Open-ended & \cmark & \xmark & \cmark \\
& MC-Bench~\cite{xu2025mc} & - & 1.5K & TI, BBox & Multi-image grounding & Referring & \xmark & \cmark & \cmark \\
\midrule

\multirow{8}{*}{Gen-S}
& VSR~\cite{liu2023vsr} & 7.6K & 3.2K & TI, TA & 2D spatial relations & T/F & \cmark & \xmark & \xmark \\
& BLINK~\cite{fu2024blink} & - & 3.9K & TI, TA, VP & Depth \& correspondence & MCQ & \cmark & \cmark & \cmark \\
& SpatialRGPT-Bench~\cite{cheng2024spatialrgpt} & - & 1.4K & TI, TA, BBox & Metric \& multi-hop logic & Open / MCQ & \cmark & \cmark & \xmark \\
& VSI-Bench~\cite{yang2025thinking} & - & 5K & TI, TA & Spatiotemporal memory & MCQ / Num & \cmark & \xmark & \cmark \\
& 3DSRBench~\cite{ma20253dsrbench} & - & 2.7K & TI, TA & 3D spatial relations & MCQ & \cmark & \xmark & \cmark \\
& SIBench~\cite{yu2025far} & - & 8.8K & TI, TA & Embodied spatial planning & MCQ / Num & \cmark & \xmark & \cmark \\
& OmniSpatial~\cite{jia2025omnispatial} & 6.9K & 1.5K & TI, TA & Dynamic \& geometric logic & MCQ / T/F & \cmark & \xmark & \cmark \\
& MINDCUBE~\cite{yin2025spatial} & 10K & 1.5K & TI, TA & Spatial mental modeling & MCQ & \cmark & \xmark & \cmark \\

\midrule

\textbf{Ours} & \textbf{CVSBench} & \textbf{19.1K} & \textbf{21.2K} & \textbf{TI, TA, BBox} & \textbf{Spatial reasoning \& grounding} & \textbf{MCQ / T/F / BBox} & \textbf{\cmark} & \textbf{\cmark} & \textbf{\cmark} \\
\bottomrule
\end{tabular}%
}
\vspace{-3mm}
\end{table*}

\subsection{CVSBench Benchmark}

Building upon the annotation pipeline described in Section~3.1, we construct CVSBench, a benchmark for evaluating cross-view spatial reasoning under controlled input protocols.  
As shown in Fig.~\ref{fig:cvsbench}, the benchmark integrates CVUSA-subsets and FOV-subsets and it  contains 3,297 image groups, 9,468 annotated BBoxes, and 40,679 QA pairs, divided into training and test sets in a 1:1 ratio.

\paragraph{Task Composition.}
CVSBench comprises several subtask types: Ground-to-Satellite (G2S), Satellite-to-Ground (S2G), Cross-View Grounding (grounding), and Viewpoint Localization (View-Arrow, View-Image).
G2S and S2G follow a single-view input protocol, requiring the model to infer spatial attributes of the unseen view. Both four-choice and binary-choice questions are included. Grounding and viewpoint localization assess cross-view entity alignment and camera pose reasoning. 

Compared with existing spatial reasoning benchmarks shown in Tab.~\ref{tab:dataset_comparison}, CVSBench differs in both task design and capability coverage. RS benchmarks focus on object detection or attribute recognition without cross-view correspondence. General Domain (Gen) benchmarks emphasize language and logical reasoning but lack geometric alignment supervision. General spatial (Gen-S) datasets primarily focus on spatial reasoning within a single image (\textit{e.g.}, 2D/3D geometric relations and depth ordering), or only address basic viewpoint transformation tasks where the differences across viewpoints are relatively small.
In contrast, CVSBench unifies cross-view VQA, grounding, and viewpoint localization within a single framework. It provides explicit bounding-box supervision, manually annotated camera pose priors, and multi-view symmetry evaluation. As shown in Tab.~\ref{tab:dataset_comparison}, CVSBench is among the few benchmarks that simultaneously support VQA, grounding, and multi-view settings (cross-view localization), and also demonstrates strong competitiveness in terms of dataset scale.

\subsection{Improving Visual Spatial Understanding Abilities}

\subsubsection{Textual Chain-of-Thought Reasoning}

Given a pre-trained multimodal large language model (MLLM), most prior work adopts a two-stage training paradigm that first performs SFT on annotated CoT data, followed by RL–based optimization. Following this paradigm, we explore two complementary textual CoT strategies with the goal of enhancing the model’s spatial reasoning capability and conduct dedicated data annotation, SFT, and RL training for each of them.

\begin{figure}[t]
    \centering
    \includegraphics[width=1.0\linewidth]{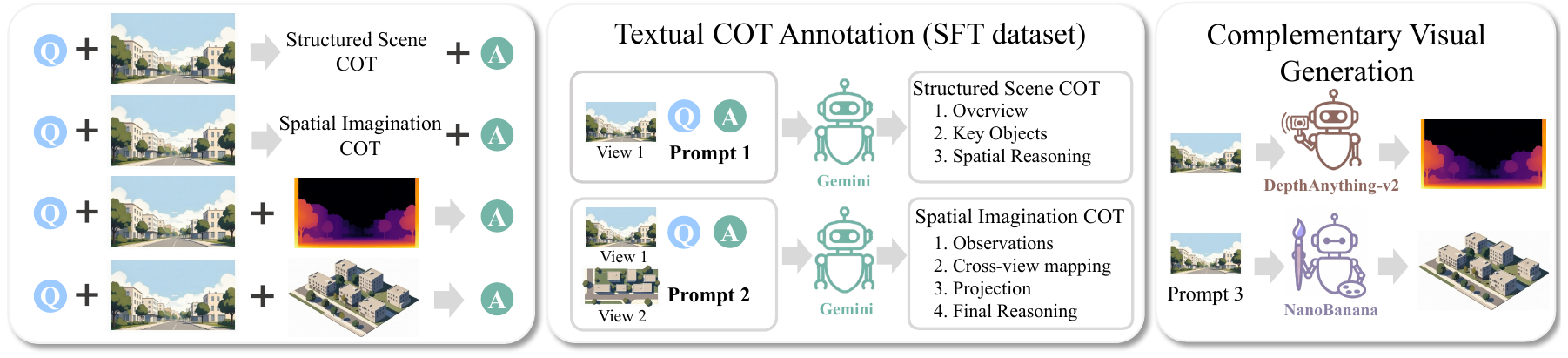} 
    \vspace{-6mm}
\caption{
Framework for enhancing spatial reasoning in VLMs. The center shows the SFT dataset annotation pipeline for textual CoT reasoning, while the right illustrates the generation of complementary visual information to support explicit spatial imagination.
}
    \label{fig:improving}
     \vspace{-4mm}
\end{figure}

\paragraph{Structured CoT.}
As shown in Fig.~\ref{fig:improving}, we begin by converting visual scenes into structured textual descriptions to facilitate the extraction of fine-grained spatial information. In the process of reasoning, the model is encouraged to explicitly identify and localize objects that are relevant to the final answer. Specifically, the model first outputs the object categories and their corresponding BBoxes, and then performs a lightweight reasoning process over the structured scene representation to derive the answer. To construct the SFT dataset, we provide the input image and the corresponding QA pair to \textsc{Gemini}~\cite{comanici2025gemini}, and prompt it to generate step-by-step structured CoT annotations that include object enumeration, spatial localization, and intermediate reasoning.

\paragraph{Imagination CoT.}
To better align with the nature of cross-view spatial understanding, we further propose a cross-view imagination CoT strategy. This approach encourages the model to mentally project object layouts observed from the input viewpoint into an alternative viewpoint and perform reasoning accordingly. During SFT data annotation, Gemini2.5-flash~\cite{comanici2025gemini} is provided with images from both viewpoints together with the QA pair, and is instructed to generate CoT annotations based on the ground-truth object distribution in the target viewpoint. This supervision enables the model to learn how spatial configurations transform across viewpoints through textual reasoning.

\paragraph{Training with GRPO.}
Based on the annotated datasets, we employ standard SFT followed by RL using the Group Relative Policy Optimization (GRPO)~\cite{shao2024deepseekmath} framework. The GRPO objective is defined as:

\begin{align}
\mathcal{J}_{\text{GRPO}}(\theta)
&= \mathbb{E}\Big[
q \sim P_{\text{sft}}(Q),
\{o_i\}_{i=1}^{G} \sim \pi_{\theta_{\text{old}}}(O \mid q)
\Big] \nonumber\\
&\quad \frac{1}{G} \sum_{i=1}^{G} \frac{1}{|o_i|}
\sum_{t=1}^{|o_i|}
\Bigg[
\hat{A}^{*}_{i,t}
- \gamma \, \mathbb{D}_{\text{KL}}\!\left(
\pi_{\theta} \,\|\, \pi_{\theta_{\text{ref}}}
\right)
\Bigg],
\end{align}
where $G$ denotes the number of sampled output sequences, $\hat{A}^{*}_{i,t}$ represents the normalized advantage at token $t$, and $\gamma$ controls the strength of KL regularization.
The KL divergence term is computed as:
\begin{align}
\mathbb{D}_{\text{KL}}[\pi_{\theta} \,\|\, \pi_{\theta_{\text{ref}}}]
=&\;
\frac{\pi_{\theta_{\text{ref}}}(o_{i,t} \mid q, o_{i,<t})}
     {\pi_{\theta}(o_{i,t} \mid q, o_{i,<t})}
 - \log
\frac{\pi_{\theta_{\text{ref}}}(o_{i,t} \mid q, o_{i,<t})}
     {\pi_{\theta}(o_{i,t} \mid q, o_{i,<t})}
 - 1.
\end{align}

The reward function is computed as a weighted combination of option-matching accuracy (0.9) and output format compliance (0.1), thereby encouraging the model to produce accurate predictions through reasoning. Additional implementation details and reward specifications are provided in the appendix.

\subsubsection{Explicit Spatial Imagination}

Purely textual descriptions and reasoning are often insufficient to fully represent complex visual scenes. When reasoning about cross-view scenarios, humans can typically construct an ``imagined world'' directly in the mind and answer questions without relying on explicit textual reasoning~\cite{newcombe2014thinking, lee2012components, bednarz2019improves}. To better simulate this human imagination process, we introduce an image generation model to construct an explicit 3D scene representation analogous to human mental imagery. We adopt nanobanana~\cite{comanici2025gemini} which is capable of understanding complex instruction to generate a 3D-view image conditioned on the input image. The generated image integrates both side-view information and top-down perspectives, thereby compensating for the inherent incompleteness of visual cues in single-view inputs.
In addition, we use a depth estimation model~\cite{yang2024depth} to generate depth maps as the additional input, in order to investigate whether introducing depth information alone, rather than cross-view visual information, can yield similar performance gains. Further details are provided in the Appendix.

\section{Experiment}
\subsection{Experimental Setup}

\paragraph{Evaluation Metrics.}
For multiple-choice tasks, we report accuracy as the proportion of correct answer, while for grounding tasks, we evaluate localization performance using mean Intersection over Union (mIoU) between predicted and ground-truth BBoxes; all results are reported as percentages.

\paragraph{Training Protocol.}
To analyze reasoning mechanisms, we use Qwen3-VL-4B~\cite{bai2025qwen3} as our base model for controlled training. The training follows two stages: SFT and RL. Half of the training data is used for SFT with Structured Scene CoT or Spatial Imagination CoT. The temperature for sampling is set to 0.01. The left portion of the dataset is used for RL training, with the temperature fixed at 0.7. Learning rate is set to 3e-5 in the SFT stage and 1e-6 in the RL stage.

\begin{table*}[t]
\centering
\small
\caption{Overall comparison on CVSBench. 
"Acc" means the overall accuracy (\%) for VQA and viewpoint tasks, and mIoU (\%) for cross-view grounding.}
\vspace{-2mm}
\scalebox{0.65}{
\setlength{\tabcolsep}{4pt}
\begin{tabular}{lcccccccccc}
\toprule
 & \multicolumn{3}{c|}{CVUSA-subset} 
 & \multicolumn{5}{c|}{FOV-subset} 
 & \multicolumn{2}{c}{Overall} \\
\cmidrule(lr){2-4}\cmidrule(lr){5-9}\cmidrule(lr){10-11}
Method 
& G2S & S2G & Grounding
& G2S & S2G & Grounding
& View-Arrow & View-Image
& Acc & mIoU \\
\midrule

human 
& 88.70 & 87.53 & 92.30
& 85.68 & 87.20 & 93.33
& 98.78 & 98.34
& 89.52 & 93.69 \\

\rowcolor{gray!10}\textbf{Close-source Models} 
& & & & & & & & & & \\

gpt-5-chat \cite{singh2025openai}
& 69.70 & 59.50 & 10.72
& 53.50 & 54.10 & 13.65
& 40.10 & 31.70
& 53.76 & 12.30\\

gpt-4o \cite{hurst2024gpt}
& 51.80 & 62.10 & 10.80
& 36.00 & 46.80 & 13.61
& 30.00 & 26.60
& 45.54 & 12.30\\

\rowcolor{gray!10}\textbf{Open-source Models} 
& & & & & & & & & & \\

Deepseekv3.2 \cite{deepseekai2025deepseekv32pushingfrontieropen}
& 63.39 & 47.03 & 3.29
& 56.69 & 38.68 & 10.98
& 26.13 & 23.92
& 44.43 & 7.39 \\

LLaVA-OV-1.5-4B-Instruct \cite{LLaVA-OneVision-1.5}
& 68.98 & 60.09 & 9.61
& 21.32 & 41.19 & 8.56
& 27.95 & 24.22
& 41.51 & 9.06 \\

Gemma-3-4b-it \cite{gemma_2025}
& 71.00 & 52.88 & 0.38
& 40.51 & 47.99 & 6.94
& 30.13 & 28.68
& 46.43 & 3.87 \\

InternVL3.5-4B \cite{wang2025internvl3}
& 58.10 & 56.00 & 2.22
& 34.30 & 44.80 & 5.53
& 39.30 & 25.70
& 44.28 & 3.99 \\

InternVL3.5-8B \cite{wang2025internvl3}
& 65.20 & 61.90 & 2.59
& 52.30 & 41.50 & 4.69
& 40.10 & 26.80
& 49.89 & 3.71 \\

Qwen3-VL-8B-Instruct \cite{bai2025qwen3}
& 69.90 & 61.90 & 0.60
& 24.00 & 46.70 & 8.10
& 42.30 & 27.50
& 45.88 & 4.59 \\

Qwen3-VL-4B-Instruct \cite{bai2025qwen3}
& 67.90 & 62.40 & 3.68
& 26.60 & 44.20 & 13.15
& 40.80 & 27.10
& 45.48 & 8.72 \\

Qwen2.5-VL-3B \cite{bai2025qwen25vltechnicalreport}
& 44.90 & 55.20 & 6.18
& 19.00 & 35.80 & 13.46
& 30.60 & 23.60
& 36.48 & 10.06 \\

Geochat-7B \cite{GeoChat2024}
& 60.69 & 56.12 & 0.30
& 56.19 & 44.44 & 0.25
& 25.59 & 28.68
& 48.23 & 0.27\\

\rowcolor{gray!10}\textbf{Spatial Reasoning Models} 
& & & & & & & & & & \\

SpaceQwen2.5VL-3B \cite{chen2024spatialvlm}
& 22.70 & 40.20 & 2.22
& 3.40 & 29.40 & 8.50
& 26.40 & 26.70
& 25.45 & 5.56 \\

SpaceThinker-Qwen2.5VL-3B \cite{chen2024spatialvlm}
& 21.80 & 34.10 & 5.46
& 19.60 & 25.30 & 8.45
& 32.40 & 23.40
& 26.66 & 7.05 \\

SpatialBot-3B \cite{SpatialBot2024}
& 68.30 & 59.08 & 0.00
& 29.18 & 42.38 & 0.02
& 26.50 & 28.68
& 43.35 & 0.01\\

ViLaSR-7B \cite{ViLaSR2024}
& 68.20 & 85.20 & 1.28
& 35.14 & 41.79 & 12.01
& 25.41 & 27.49
& 44.05 & 6.39\\

\bottomrule
\end{tabular}}
\vspace{-2mm}
\label{tab:main}
\end{table*}

\subsection{Main Benchmark Results}
\paragraph{Human Baseline.}
To calibrate task difficulty, we collect human performance on the test set. 
The questions are presented through a unified web-based UI, and eight persons independently provide answers. 
We report the aggregated accuracy across annotators. 
As shown in Tab.~\ref{tab:main}, human performance substantially exceeds current VLMs, indicating that the benchmark remains solvable while being challenging for existing models.

\paragraph{Model-Level Comparison.}
Tab.~\ref{tab:main} presents the overall performance on CVSBench. Closed-source models achieve the strongest results, with gpt-5-chat ranking highest. Among open-source systems, InternVL3.5-8B~\cite{wang2025internvl3} performs best and approaches closed-source performance on VQA tasks. In contrast, spatial reasoning–oriented models (\textit{e.g.} , SpaceQwen~\cite{chen2024spatialvlm} and SpaceThinker~\cite{chen2024spatialvlm}) underperform general-purpose VLMs overall. Although these models are designed to enhance single-view spatial understanding through stronger geometric and depth-aware reasoning, such improvements do not directly translate into better cross-view correspondence. Moreover, their ability to recognize attributes that rely on fine-grained texture details cues such as facades and materials is relatively weak. Consequently, these models remain limited in tasks that require maintaining structural and semantic consistency across multiple viewpoints.

\paragraph{Task-Level Difficulty.}
Cross-view VQA tasks (G2S and S2G) exhibit moderate performance, suggesting that coarse-grained cross-view associations are partially solvable. In contrast, grounding tasks consistently achieve low IoU scores, indicating insufficient capability in establishing cross-view object correspondence. This limitation can be attributed to two factors: the lack of training on multi-image grounding scenarios and the substantial appearance variation of the same object across different viewpoints. As a result, models struggle to  achieve precise entity-level alignment across views. Furthermore, performance under the FOV-subset is consistently lower than that on CVUSA-subset, demonstrating that a restricted field of view and incomplete structural cues further exacerbate the difficulty of cross-view correspondence.

\paragraph{View Alignment Phenomenon.}
In the Viewpoint Localization task, arrow selection is consistently easier than image selection. The former involves matching multiple arrows in a RS image with a single street-view image, whereas the latter requires modeling the correspondence between four candidate street-view images and a single arrow in the RS image. Consequently, image selection demands processing a larger amount of visual information to extract feature correspondences. This performance gap highlights the limitations of VLMs in handling cross-view visual feature representation and entity-level alignment.

\begin{table*}[t]
\centering
\caption{Effect of inference-time CoT prompting on Qwen3-VL-4B~\cite{bai2025qwen3}. The model is required to first output reasoning traces before producing the final answer.}
\vspace{-2mm}
\small
\setlength{\tabcolsep}{4pt}
\resizebox{\textwidth}{!}{
\begin{tabular}{lccc|ccccc}
\toprule
 & \multicolumn{3}{c|}{CVUSA-subset} 
 & \multicolumn{5}{c}{FOV-subset} \\
\cmidrule(lr){2-4}\cmidrule(lr){5-9}
CoT
& G2S & S2G & Grounding
& G2S & S2G & Grounding
& View-Arrow & View-Image \\
\midrule

None
& 67.90 & 62.40 & 3.68
& 26.70 & 43.90 & 13.15
& 40.80 & 27.10 \\

Inference CoT
& 69.20 & 61.70 & 6.01
& 35.60 & 45.30 & 12.39
& 21.90 & 23.60 \\

\bottomrule
\end{tabular}}
\vspace{-3mm}
\label{tab:inference_cot}
\end{table*}

\subsection{Textual Reasoning Exploration}

\begin{table*}[t]
\centering
\caption{Performance comparison (\%) of different training strategies and CoT designs on  G2S and S2G of CVUSA-subset. Top two results are highlighted in green: dark green for the first and light green for the second. Footprint: Building footprint area; Connection: Building connection; Distance: Distance to camera; Roof: Roof form.}
\vspace{-2mm}
\small
\scalebox{0.66}{
\setlength{\tabcolsep}{6pt}
\begin{tabular}{lc|cccc|cccc}
\toprule
  &  & \multicolumn{4}{c|}{ CVUSA-subset G2S} & \multicolumn{4}{c}{ CVUSA-subset S2G} \\
\cmidrule(lr){3-6}\cmidrule(lr){7-10}
 Training & CoT
& Footprint
& Connection
& Distance 
& Roof
& Height
& Location
& Vegetation
& Visibility \\
\midrule

None
& None
& 62.8 & 25.0 & 84.0 & \cellcolor{bestgreen}63.2
& \cellcolor{bestgreen}59.5 & 64.0 & \cellcolor{bestgreen}77.4 & 47.9 \\

SFT
& Structured Scene
& \cellcolor{secondgreen}63.2 & 28.1 & 88.7 & 50.5
& \cellcolor{secondgreen}53.3 & 63.4 & \cellcolor{secondgreen}74.8 & \cellcolor{bestgreen}57.5 \\

+ RL
& Structured Scene
& 58.6 & \cellcolor{bestgreen}37.2 & 87.5 & 49.2
& \cellcolor{secondgreen}53.3 & \cellcolor{bestgreen}65.1 & 73.1 & \cellcolor{secondgreen}55.9 \\

SFT
& Spatial Imagination
& \cellcolor{bestgreen}64.6 & \cellcolor{secondgreen}31.3 & \cellcolor{secondgreen}89.1 & 46.5
& 50.3 & 62.0 & 73.9 & 52.4 \\

+ RL
& Spatial Imagination
& \cellcolor{secondgreen}63.2 & 31.2 & \cellcolor{bestgreen}89.8 & \cellcolor{secondgreen}62.0
& 48.7 & \cellcolor{secondgreen}64.6 & 74.6 & 53.3 \\

\bottomrule
\end{tabular}}
\vspace{-4mm}
\label{tab:cvusa_cot}
\end{table*}

\begin{table*}[t]
\centering
\caption{Performance comparison (\%) of different training strategies and CoT designs on G2S and S2G of FOV-subset. Color: Facade color; Material: Ground material; Height: Height comparison; Occlusion: Occlusion binary; Position: Relative position.}
\vspace{-2mm}
\small
\scalebox{0.61}{
\setlength{\tabcolsep}{6pt}
\begin{tabular}{lc|ccc|cccccc}
\toprule
  &  & \multicolumn{3}{c|}{FOV-subset G2S} & \multicolumn{6}{c}{FOV-subset S2G} \\
\cmidrule(lr){3-5}\cmidrule(lr){6-11}
 Training & CoT
& Facade
& Roof
& Symmetry
& Color
& Material
& Height
& Occlusion
& Position
& Vegetation Sector \\
\midrule
 None
& None
& 18.4 & 9.9 & \cellcolor{bestgreen}79.8
& 24.5 & 41.5 & \cellcolor{bestgreen}73.6 & \cellcolor{secondgreen}51.9 & \cellcolor{secondgreen}56.1 & 49.0 \\

 SFT
& Structured Scene 
& 41.3 & \cellcolor{bestgreen}14.8 & \cellcolor{secondgreen}72.9
& \cellcolor{bestgreen}32.0 & \cellcolor{bestgreen}46.2 & 65.5 & 48.1 & \cellcolor{secondgreen}56.1 & \cellcolor{secondgreen}51.6 \\

  + RL
& Structured Scene
& 25.5 & 11.2 & 56.0
& \cellcolor{secondgreen}29.0 & \cellcolor{secondgreen}45.4 & 64.2 & 42.4 & \cellcolor{bestgreen}58.7 & \cellcolor{bestgreen}51.7 \\

SFT
&  Spatial Imagination
& \cellcolor{bestgreen}51.0 & \cellcolor{secondgreen}14.1 & 57.1
& 27.7 & 43.8 & 66.7 & \cellcolor{bestgreen}54.8 & 55.8 & 49.5 \\

  + RL
&  Spatial Imagination
& \cellcolor{secondgreen}43.3 & 8.4 & 67.5
& 27.0 & 43.9 & \cellcolor{secondgreen}69.6 & 42.4 & 55.9 & 51.0 \\
\bottomrule
\end{tabular}}
\vspace{-3mm}
\label{tab:fov_cot}
\end{table*}

We investigate whether modifying textual reasoning improves cross-view spatial understanding. 
We prompt the model to reason at inference time without further training. As shown in Tab.~\ref{tab:inference_cot}, inference-time CoT yields marginal gains in certain categories, particularly grounding and FOV-subset G2S. Grounding under CVUSA shows slight improvement, suggesting that reasoning can partially guide spatial correspondence.
However, improvements are inconsistent across different tasks, indicating that expanding reasoning during inference alone cannot fundamentally improve cross-view alignment.
We further train and evaluate models under two supervised reasoning formats: Cross-View Imagination CoT and Structured Scene CoT, where each format is optimized through SFT followed by RL.
These augmentation strategies are applied only to cross-view VQA tasks, and not to grounding or viewpoint localization.

\begin{table*}[t]
\centering
\caption{Performance comparison (\%) of different auxiliary views for Qwen3VL-4B on the FOV-subset G2S and S2G tasks. 
}
\vspace{-2mm}
\small
\scalebox{0.68}{
\setlength{\tabcolsep}{6pt}
\begin{tabular}{l|ccc|cccccc}
\toprule
 & \multicolumn{3}{c|}{FOV-subset G2S} & \multicolumn{6}{c}{FOV-subset S2G} \\
\cmidrule(lr){2-4}\cmidrule(lr){5-10}
Auxiliary View
& Facade
& Roof
& Symmetry
& Color
& Material
& Height
& Occlusion
&  Position
& Vegetation Sector \\
\midrule

None
& \cellcolor{secondgreen}18.4 & 9.9 & \cellcolor{secondgreen}79.8
& \cellcolor{secondgreen}24.5 & 41.5 & 73.6 & \cellcolor{secondgreen}51.9 & \cellcolor{secondgreen}56.1 & \cellcolor{secondgreen}49.0 \\

Depth
& \cellcolor{bestgreen}18.5 & \cellcolor{secondgreen}10.2 & \cellcolor{secondgreen}79.8
& 24.4 & \cellcolor{secondgreen}42.2 & \cellcolor{secondgreen}74.3 & \cellcolor{bestgreen}54.8 & \cellcolor{secondgreen}56.1 & 48.8 \\

3D View
& 17.1 & \cellcolor{bestgreen}18.7 & \cellcolor{bestgreen}80.5
& \cellcolor{bestgreen}36.9 & \cellcolor{bestgreen}47.0 & \cellcolor{bestgreen}75.0 & 51.2 & \cellcolor{bestgreen}58.3 & \cellcolor{bestgreen}50.0 \\
\bottomrule
\end{tabular}}
\vspace{-4mm}
\label{tab:banana}
\end{table*}

\begin{figure}[h]
    \centering
    \includegraphics[width=1.0\linewidth]{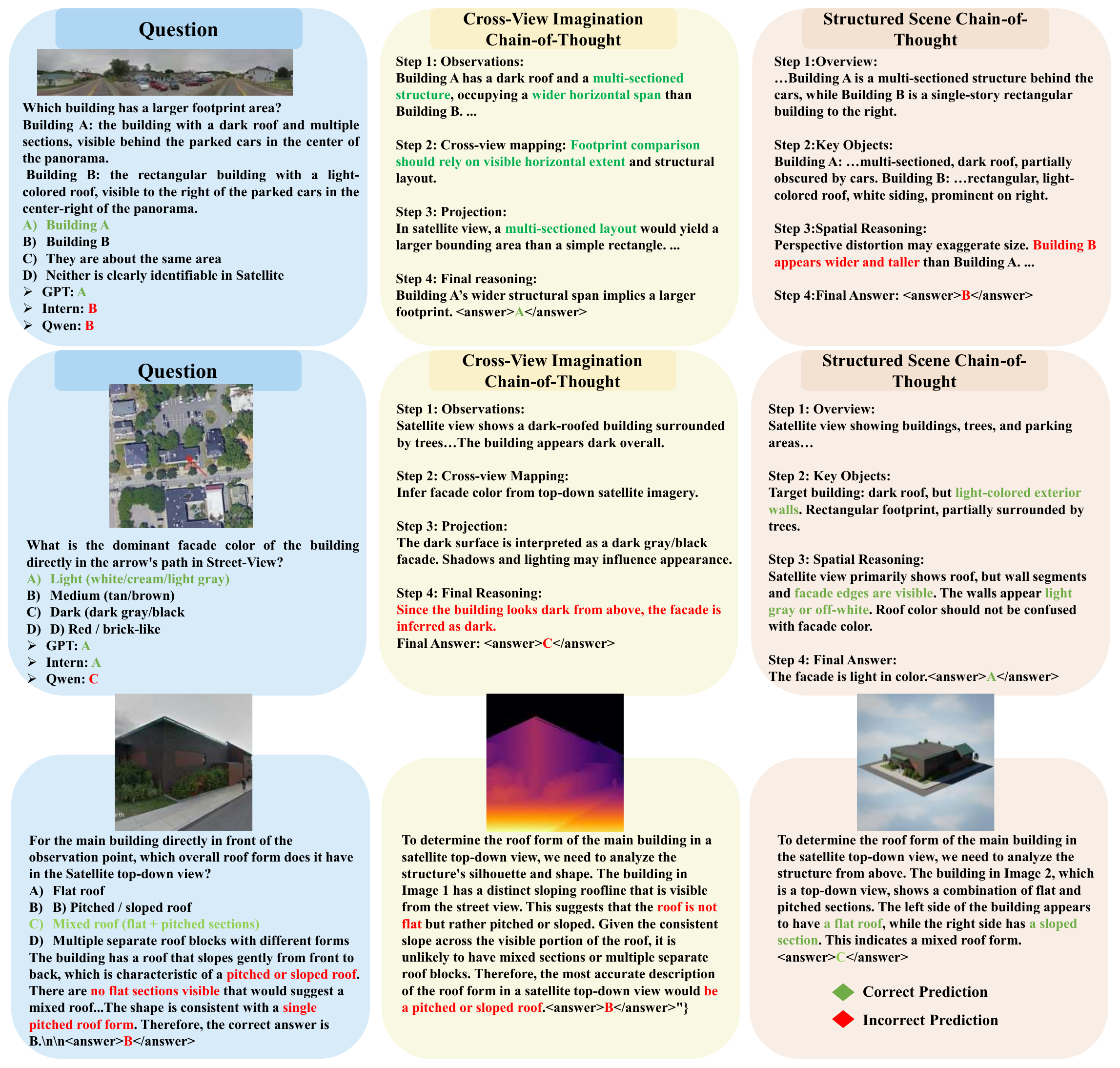}
    \vspace{-4mm}
    \caption{\textbf{Qualitative examples.} A comparison of methods for enhancing the spatial understanding capabilities of VLMs.  \textcolor{green}{Green} indicates GT or correct prediction. \textcolor{red}{Red} indicates incorrect prediction.}
    \label{fig:qual_case}
    \vspace{-4mm}
\end{figure}

\paragraph{Structured CoT.}
Structured CoT enforces object-centric decomposition by grounding entities before attribute reasoning. As shown in Tab.~\ref{tab:cvusa_cot} and Tab.~\ref{tab:fov_cot}, it improves several attributes that require cross-view entity correspondence and spatial alignment. These persistent improvements suggest that grounding-first decomposition provides a relatively stable reasoning scaffold rather than transient gains driven by superficial cues. Fig.~\ref{fig:qual_case} shows representative cases. For example, it separates roof regions from facade regions when predicting facade color from satellite views, and distinguishes perspective effects when comparing building footprints across views. By disentangling regions before attribute prediction, it mitigates cross-view attribute confusion and leads to more stable improvements.

\paragraph{Imagination CoT.}
This paradigm introduces an explicit projection process from ground view to a hypothesized satellite representation. On CVUSA-subset, it improves geometry-sensitive categories such as distance estimation, suggesting better approximation of latent 3D correspondence. 
As shown in Fig.~\ref{fig:qual_case}, imagination-style projection can correctly reason about footprint span, whereas Structured Scene COT primarily focuses on the scene layout of the input viewpoint and its visual cues are more susceptible to perspective-induced bias.
However, improvements on appearance-dominant attributes are less consistent across datasets. These improvements are category-specific and remain limited on FOV-subset. 

Additionally, in the FOV-subset, performance on certain types drops significantly after RL compared to SFT. This is primarily because these tasks rely on fine-grained visual cues that are difficult to accurately resolve through textual reasoning alone (\textit{e.g.}, facade color types depend on boundary color, roof types require attention to roof inclination, and occlusion types requires projection along the observation viewpoint). Consequently, the model struggles to learn answering strategies through reasoning, and after RL training it exhibits preferences toward specific answer options during evaluation, resulting in reward hacking.

\subsection{Explicit Spatial Representation Exploration}

We investigate whether strengthening spatial representation at the input level leads to more stable improvements. 
We apply the 3D‑miniature and depth augmentations only to the FOV-subset because FOV provides limited visual cues and benefits more from additional 3D information, whereas CVUSA-subset already contains richer panoramic context and our experiments show that it is difficult to reliably generate 3D-view images, due to the limited capability of existing image generation models in understanding panoramic imagery.

\paragraph{Depth Augmentation.}

Depth augmentation primarily enriches local geometric ordering within the same viewpoint. As shown in Tab.~\ref{tab:banana}, depth augmentation mainly improves geometry-sensitive categories such as occlusion and height under FOV-subset. By making foreground–background relations explicit, it improves geometry-sensitive categories such as occlusion and height in FOV-subset. However, gains on appearance-dependent attributes (e.g., facade and material) remain limited. This suggests that enhancing single-view depth cues refines local geometry but does not substantially improve cross-view structural alignment.

\paragraph{3D Miniature Rendering.}

In contrast, the 3D miniature view introduces a more explicit global structural representation. As shown in Tab.~\ref{tab:banana}, the 3D view yields broader gains in facade color recognition, position estimation, and roof-related categories. Rather than refining local geometry, it provides a compact structural abstraction closer to satellite perspective. These results indicate that modeling global structural layout is more effective for cross-view correspondence than strengthening single-view geometric detail alone. As shown in Fig.~\ref{fig:qual_case}, single-view cues may mislead global roof inference. 
Depth improves local geometry, whereas the 3D view reveals global structure, explaining its stronger gains over both depth and text-only CoT.
More details about generation and implementation are provided in the Appendix.

\section{Conclusion}
We present CVSBench, a large-scale benchmark for evaluating cross-view spatial reasoning and imagination in VLMs under the satellite–street view setting. Our study reveals that current VLMs struggle to maintain consistent spatial representations when facing large viewpoint changes, highlighting a fundamental limitation beyond static spatial understanding. We explore several reasoning strategies and find that purely text-based approaches offer limited improvements, while incorporating visual spatial imagination shows greater potential for enhancing cross-view reasoning. These findings suggest that future VLMs should move beyond language-only reasoning and integrate explicit perceptual imagination mechanisms. 
We hope CVSBench will facilitate further research toward more robust and human-like spatial cognition in VLMs.

\section*{Acknowledgements}
This work is supported by Fundamental and Interdisciplinary Disciplines Breakthrough Plan of the Ministry of Education of China (No. JYB2025XDXM101), China NSFC Projects under Contract 62272375 and Tianyuan Fund for Mathematics of the National Natural Science Foundation of China (Grant No. 12426105).


%
%
\bibliographystyle{splncs04}
\bibliography{main}
\newpage

\appendix

\section{Dataset Construction and Analysis}

\subsection{Annotation Interface}
\begin{figure}[htbp]
    \centering
    \begin{subfigure}[b]{0.31\textwidth}
        \centering
        \includegraphics[width=\textwidth]{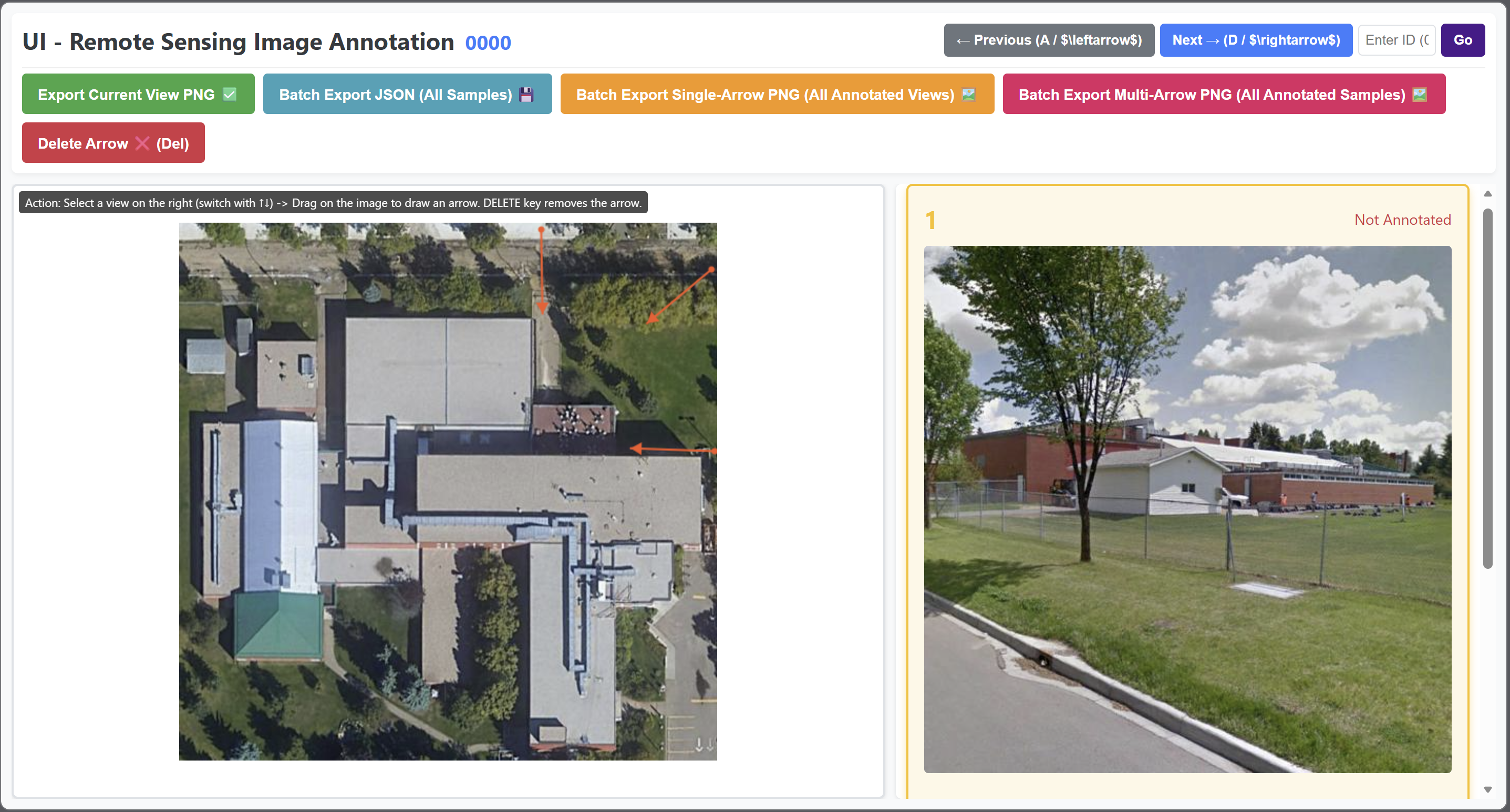}
        \caption{}
        \label{fig:html}
    \end{subfigure}
    \hfill
    \begin{subfigure}[b]{0.31\textwidth}
        \centering
        \includegraphics[width=\textwidth]{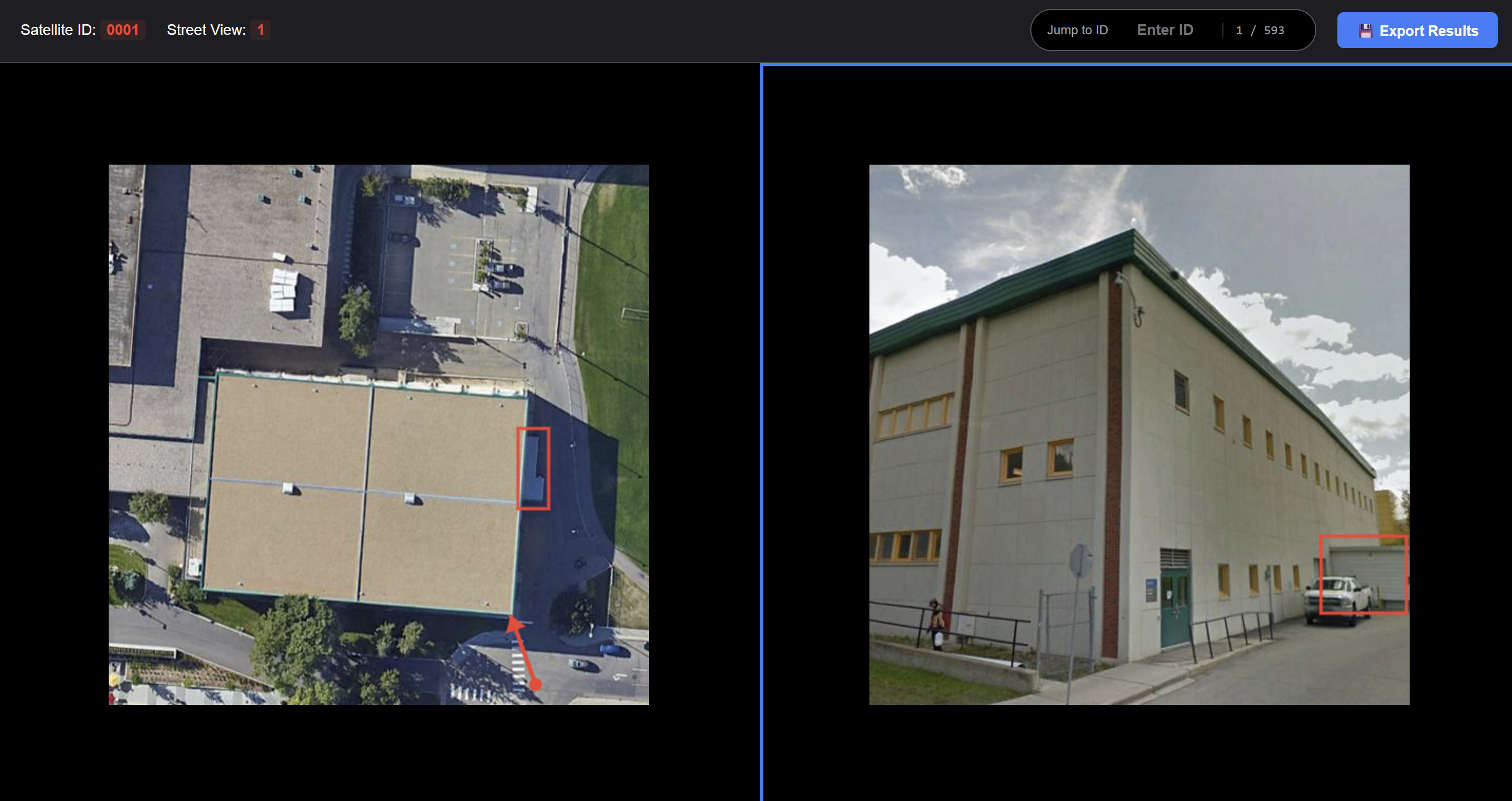}
        \caption{}
        \label{fig:html2}
    \end{subfigure}
    \hfill
    \begin{subfigure}[b]{0.31\textwidth}
        \centering
        \includegraphics[width=\textwidth]{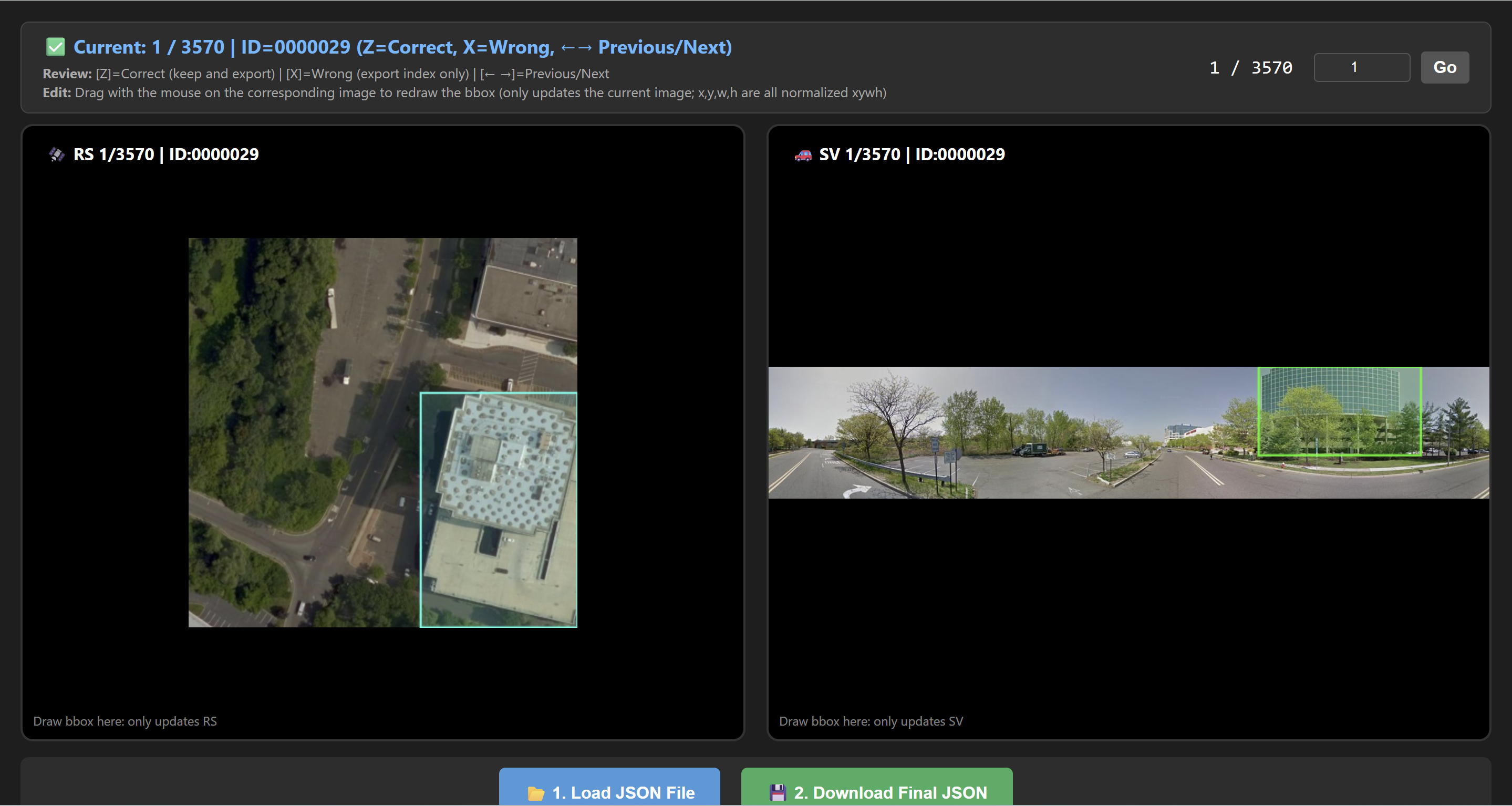}
        \caption{}
        \label{fig:html3} 
    \end{subfigure}

    \caption{
    Task-specific annotation interfaces used in dataset construction.
    (a) Viewpoint annotation interface for the FOV-subset, where annotators label
    camera locations and viewing directions using arrows on the satellite image.
    (b) Cross-view bounding-box annotation interface for the FOV-subset, built upon
    the previously annotated viewpoint arrows.
    (c) Bounding-box verification interface for the CVUSA-subset, where
    Gemini-generated annotations are manually checked and corrected.
    }
    \label{fig:annotation_interface}
\end{figure}
To support the construction of CVSBench, we developed three dedicated
web-based annotation interfaces for different annotation stages and data
sources. These interfaces correspond to: (1) viewpoint annotation for the
FOV-subset, (2) object bounding-box annotation for the FOV-subset based on
previously annotated viewpoints, and (3) bounding-box verification for the
CVUSA-subset based on Gemini-generated initial annotations.

As shown in Fig.~\ref{fig:annotation_interface}(\subref{fig:html})
 the interface is used for viewpoint annotation in the FOV-subset. Annotators mark directional
arrows on the satellite image, where the arrow origin indicates the camera
location and the arrow direction denotes the viewing orientation. These arrow
annotations provide the geometric prior for subsequent viewpoint localization
tasks and also serve as the basis for later cross-view object annotation.

As shown in Fig.~\ref{fig:annotation_interface}(\subref{fig:html2}), the interface is used for object-level bounding-box annotation in the FOV-subset. Based on the previously annotated camera arrows, annotators
inspect the satellite image and the corresponding street-view image together,
and manually draw bounding boxes for corresponding objects across views. This
interface is designed to facilitate accurate alignment between the limited-view
ground image and the satellite image under large viewpoint differences.

As shown in Fig.~\ref{fig:annotation_interface}(\subref{fig:html3}), the interface is used for the CVUSA-subset. In this subset, initial
object bounding boxes are first generated automatically by Gemini. Annotators
then use the verification interface to review, adjust, and correct these
pre-annotated results. This process focuses on improving boundary accuracy,
cross-view consistency, and annotation completeness, while reducing the manual
cost of annotating panoramic cross-view correspondences from scratch.

Overall, the annotation pipeline is semi-automatic. 
Viewpoint arrows, object bounding boxes, and Gemini-generated candidates are refined by human annotators through these task-specific interfaces. This process ensures high-quality data for viewpoint localization, cross-view grounding, and downstream QA construction.

In the CVUSA-subset, street-view panoramas have a resolution of 1232$\times$224
and satellite images have a resolution of 370$\times$370. In the FOV-subset,
street-view images are stored at 512$\times$512, while satellite images are
stored at 512$\times$512.

The resulting annotations are stored in a lightweight JSON format.
For viewpoint annotations, each entry 
specifies
the sample ID, the street-view
index, the normalized camera location on the satellite image, and the viewing
direction. The camera position is represented by normalized coordinates
$(x, y)$ with respect to the satellite image, where $(0,0)$ corresponds to the
top-left corner and $(1,1)$ corresponds to the bottom-right corner.
The viewing direction is represented by an angle in degrees, where
$0^\circ$ indicates the upward direction of the satellite image with angles increasing clockwise.

An example annotation is shown below:

\begin{verbatim}
{
 "sample_id": "0001",
 "annotations": [
   {"view_id":1,"x":0.79,"y":0.96,"angle":340.2}
 ]
}
\end{verbatim}

For cross-view grounding annotations, each object is represented by a pair of
normalized bounding boxes across the satellite and
street-view images. Each bounding box is represented by $(x,y,w,h)$ where $(x, y)$ denotes the top-left corner and $(w, h)$ specifies the dimensions, with all values normalized relative to the respective image resolution.

\begin{verbatim}
{
 "0842_1": {
   "rs":[{"x":0.75,"y":0.33,"w":0.16,"h":0.10}],
   "sv":[{"x":0.19,"y":0.12,"w":0.25,"h":0.16}]
 }
}
\end{verbatim}

\subsection{Data Generation Prompts and Instructions}

For cross-view VQA generation, we use structured prompts to construct
question--answer pairs for two inverse settings: Ground-to-Satellite (G2S)
and Satellite-to-Ground (S2G). During data generation, the annotation model
is provided with \emph{both} the satellite image and the corresponding
street-view image, enabling it to verify cross-view consistency and produce
high-quality questions. During benchmark evaluation, however, the solver is
restricted to \emph{only one} of the two views depending on the task setting.
Therefore, the prompts are designed to ensure that the question is answerable
from the solver-side input, while the correctness of the answer is guaranteed
by the complementary view.

The prompt design follows three shared principles. First, each question must
require genuine cross-view reasoning rather than direct single-view reading.
Second, targets must be defined using stable and uniquely identifiable anchors,
such as building structures, road relations, or geometric layouts, while
temporary objects such as vehicles, people, and animals are prohibited.
Third, the prompts explicitly forbid answer leakage, including directional
hints, semantic building names, and descriptive words that directly imply the
correct option.

Since the CVUSA-subset and the FOV-subset have different camera geometry, we
use four prompt variants in total: G2S-CVUSA, G2S-FOV, S2G-CVUSA, and
S2G-FOV. Their shared parts are summarized above, while the task-specific
differences are given below.

\subsubsection{Ground-to-Satellite (G2S) Prompt}

In G2S tasks, the solver observes street-view input during evaluation,
while the answer is verified from the satellite image. We use two prompt
variants: one for the FOV-subset with an observation point and viewing arrow,
and one for the CVUSA-subset with panorama--satellite alignment rules.
For the FOV-subset, most questions are generated from a single street-view
image. However, for symmetry-related questions, we allow the generator to
access up to four street-view views from the same location so that the model
can better inspect the building from multiple nearby perspectives before
determining whether the target building is symmetric in satellite view.

\paragraph{G2S-FOV Prompt.}
For the FOV-subset, the default input consists of one street-view image and
one satellite image. For symmetry-related cases, the street-view input may
contain up to four views from the same sample, which provide complementary
facade observations of the target building.
\begin{tcolorbox}[breakable]
You are an expert dataset annotator creating multiple-choice questions for
Street-View-to-Satellite-View (Ground2Sat) cross-view reasoning.

You will be provided with:
- A Satellite Map (top-down view) including an observation point and
  viewing-direction arrow.
- A Street-View image captured at the SAME observation point, facing EXACTLY
  the arrow direction.

=========================================
[CRITICAL CAMERA GEOMETRY FACT]

The Street-View camera is located at the observation point in the Satellite Map.
The arrow defines the Street-View ``forward'' direction.

=========================================
[STREET-VIEW SELF-CONTAINED DIRECTIONS]

Forward = facing into the image center.
Left/Right = your left/right when facing forward.
Behind = opposite of forward.

Never use compass directions (N/E/S/W).
Never use satellite-only up/down/left/right.

=========================================
[CORE TASK --- HARD]

During question generation, you see BOTH views.
During evaluation, the solver sees ONLY the Street-View image.

Each question MUST:
- Be solvable using ONLY Street-View reasoning.
- Have a correct answer that is clearly and directly decidable in the
  Satellite top-down view.

=========================================
STRICTLY FORBIDDEN (DISCARD IMMEDIATELY)

NO vehicles (any form).
NO people or animals.
NO Street-View-only questions.
NO fantasy distractors.
NO uncertain / undecidable answers.
NO satellite-only identifiers (IDs, coordinates, map corners).
NO unstable anchors (trees, shadows, poles, small signs).
NO asking about objects not visible in Street-View.
NO numbered entrance anchors (e.g., entrance 1/2/3).

=========================================
NO ANSWER LEAKAGE (MANDATORY)

The question stem MUST NOT contain words that directly imply the correct option.

Forbidden words in stems:
- flat roof / pitched roof / mixed roof
- indentation / protrusion (as conclusion wording)
- L-shaped / rectangular / compact
- attached / merged / extension

The stem may describe visible Street-View features ONLY to locate the anchor.
It must NOT reveal the answer category.

=========================================
ANCHOR REQUIREMENTS (MANDATORY)

All anchors MUST:
- Be clearly visible in Street-View,
- Be uniquely locatable using Street-View-only wording,
- Be stable architectural structures.

Preferred anchors:
- a recessed wall bay / notch-like wall area
- a vertical block-like wall feature
- a clearly visible window band/cluster
- a rooftop box-like structure visible from Street-View
- a clearly visible roof ridge/edge line
- a clearly visible public utility/service equipment area adjacent to the
  building (ONLY if clearly visible)

=========================================
SATELLITE DECIDABLE (HARD)

For every question:
- The correct answer MUST be directly and unambiguously visible in Satellite view.
- Wrong options must be clearly false.

If Satellite cannot clearly determine the answer -> DISCARD and regenerate.

=========================================
MUST REQUIRE CROSS-VIEW REASONING

The answer must NOT be directly readable from Street-View alone.
Street-View provides clues.
Satellite provides the ground-truth.

=========================================
TASK: GENERATE EXACTLY 3 QUESTIONS (FIXED TYPES)

For EACH street+sat pair, output EXACTLY 3 questions:

---------
Q1 --- FACADE FEATURE VISIBILITY (2 OPTIONS)
---------
Choose ONE clearly visible facade-level feature in Street-View.

Examples:
- recessed wall section
- vertical wall block
- visible window band
- clearly visible public utility/service equipment area

Ask whether this feature is visible/identifiable in the Satellite top-down view.

Q1 options MUST be exactly:
A) Visible in Satellite
B) Not visible in Satellite

-----------------------
Q2 --- ROOF FORM (4 OPTIONS)
-----------------------
Question stem MUST be:

``For the main building directly in front of the observation point, which
overall roof form does it have in the Satellite top-down view?''

IMPORTANT:
- The answer MUST require Satellite confirmation.
- If roof form is directly obvious from Street-View -> DISCARD and regenerate.

Q2 options MUST be exactly:
A) Flat roof
B) Pitched / sloped roof
C) Mixed roof (flat + pitched sections)
D) Multiple separate roof blocks with different forms

--------------
Q3 --- BUILDING SYMMETRY (2 OPTIONS)
--------------
For this question type, up to four street-view views from the same FOV sample
may be provided, so that symmetry-related facade evidence can be inspected
from multiple nearby perspectives. Ask whether the main building directly in front appears left-right symmetric
in the Satellite top-down view.

This must be directly decidable from Satellite.
If symmetry cannot be clearly determined -> DISCARD.

Q3 options MUST be exactly:
A) Symmetric
B) Not symmetric

=========================================
ANSWER DISTRIBUTION (MANDATORY)

Across the 3 questions, correct answers must use at least 3 distinct letters
among \{A,B,C,D\}. (Q1 and Q3 only use A/B. Ensure Q2 introduces additional
letters.)

=========================================
OUTPUT FORMAT

Output ONLY valid JSON.
Exactly 3 objects in random order:

[
  {
    ``question'': ``...'',
    ``options'': [``A) ...'',``B) ...''],
    ``answer'': ``A'',
    ``category'': ``facade''
  },
  {
    ``question'': ``...'',
    ``options'': [``A) ...'',``B) ...'',``C) ...'',``D) ...''],
    ``answer'': ``C'',
    ``category'': ``roof''
  },
  {
    ``question'': ``...'',
    ``options'': [``A) ...'',``B) ...''],
    ``answer'': ``B'',
    ``category'': ``symmetry''
  }
]
\end{tcolorbox}

\paragraph{G2S-CVUSA Prompt.}
The CVUSA prompt shares the same core constraints as the FOV version, but the
camera geometry is different: the panorama is captured at the \emph{same
location as the center of the satellite image}, and there is \emph{no arrow}
indicating the viewing direction. Instead, the prompt defines a fixed
panorama--satellite alignment: panorama center corresponds to the top side of
the satellite map, the left quarter corresponds to the left side, the right
quarter corresponds to the right side, and the far panorama edges correspond
to the bottom side of the satellite image. In addition, the CVUSA G2S prompt
restricts the allowed categories to \texttt{distance\_to\_camera},
\texttt{building\_footprint\_area}, \texttt{building\_connec-} \texttt{tion}, and
\texttt{roof\_form}, and enforces stronger ``extreme-only'' rules for distance
and footprint-area questions to avoid ambiguous comparisons. The full prompt
is given below.

\begin{tcolorbox}[breakable]
You are an expert dataset annotator creating cross-view reasoning
multiple-choice questions.

You will be provided with:
1) A Satellite Map (top-down view).
2) A Street-View PANORAMA image (ground-level, 360-degree).

=========================================
[CRITICAL CAMERA GEOMETRY FACT --- HARD]

The panorama is captured at the SAME location as the CENTER of the Satellite Map.
The camera location is exactly at the satellite image center.

There is NO arrow and NO given viewing direction in the Satellite Map.

=========================================
[CRITICAL PANORAMA-TO-SATELLITE ALIGNMENT FACT --- HARD]

- Center region of the panorama <-> TOP side of the Satellite Map.
- Left \& right edges of the panorama <-> BOTTOM side of the Satellite Map.
- Left quarter of the panorama <-> LEFT side of the Satellite Map.
- Right quarter of the panorama <-> RIGHT side of the Satellite Map.

You MUST use this mapping to align Street-View targets to Satellite objects.

Do NOT use compass words.
Do NOT use ``forward / ahead / directly in front''.

=========================================
[CORE TASK --- HARD]

During generation, YOU can see BOTH views.
During evaluation, the solver sees ONLY the panorama.

Therefore:
- Each question MUST be solvable using ONLY the panorama.
- The correct answer MUST be 100\% guaranteed by Satellite.

ZERO-TOLERANCE RULE:
If you cannot guarantee the correct answer from Satellite with absolute
certainty -> DISCARD and regenerate.

=========================================
HARD BANS

NO VEHICLES.
NO PEOPLE / ANIMALS.
NO compass words.
NO satellite-only IDs/coordinates/building numbers.
NO ``unclear/unknown/undecidable'' options.
NO fantasy distractors.

=========================================
SATELLITE-FIRST GATING WORKFLOW (NEW --- MUST FOLLOW)

Step 1 (Satellite check):
- Identify candidate target(s) near the satellite center.
- Verify the needed relation/property is unambiguous in Satellite.
- If NOT unambiguous -> abandon that candidate immediately.

Step 2 (Street-View check):
- Confirm the SAME target(s) can be uniquely located in the panorama using
  only visible cues + panorama-position anchors.
- If NOT uniquely locatable -> abandon.

Step 3 (Only then) write the question.

=========================================
BUILDING IDENTIFICATION MUST BE IN THE QUESTION BODY (HARD)

For any building question, the question text MUST define targets clearly.

For TWO-building questions, include in the QUESTION BODY:

Building A: [Neutral Street-View unique descriptor + panorama-position anchor]
Building B: [Neutral Street-View unique descriptor + panorama-position anchor]

Options MUST NOT contain descriptions.
Options may ONLY say ``Building A / Building B'' (plus template options C/D).

=========================================
EXTREME-ONLY RULE --- distance\_to\_camera (HARD)

Distance questions are OPTIONAL.

Generate a distance question ONLY if the Satellite relation is EXTREME and
cannot be mistaken:
- Building A is very close to the satellite center AND
- Building B is much farther away (clear, large gap).

If distance is even remotely close -> DO NOT generate.

=========================================
EXTREME-ONLY RULE --- building\_footprint\_area (HARD)

Area questions are OPTIONAL.

Generate an area question ONLY if the footprint difference is EXTREME and
cannot be mistaken:
- One footprint is obviously much larger (e.g., about \textasciitilde 2$\times$
  visually or clearly dominates the other).

NO SIZE-HINT WORDS in building descriptions.

=========================================
STRICT RULE --- building\_connection (HARD)

Generate ONLY when connectivity is genuinely uncertain:
- Buildings are adjacent or nearly touching in Satellite.
- Not separated by a road or obvious gap.
- Not trivially ``No''.

=========================================
STRICT RULE --- roof\_form (HARD)

Generate ONLY if Satellite makes roof form unambiguous.

Do NOT ask: ``Which building has a gabled roof?''
Instead pick ONE clearly described building and ask:
``What roof form does this building have?''

=========================================
ALLOWED QUESTION CATEGORIES (ONLY 4)

1) ``distance\_to\_camera''
2) ``building\_footprint\_area''
3) ``building\_connection''
4) ``roof\_form''

=========================================
QUESTION COUNT

Generate EXACTLY 3 questions per image pair.

If distance/area cannot meet EXTREME-ONLY rule -> do NOT force them.
Use roof\_form and/or building\_connection instead.

=========================================
MULTIPLE-CHOICE FORMAT

Options must be exactly:
[``A) ...'', ``B) ...'', ``C) ...'', ``D) ...'']

Answer must be: ``A'' / ``B'' / ``C'' / ``D''.

=========================================
CATEGORY OPTION TEMPLATES (STRICT)

distance\_to\_camera:
A) Building A
B) Building B
C) They are about the same distance
D) Both are not visible in Satellite

building\_footprint\_area:
A) Building A
B) Building B
C) They are about the same area
D) Neither is clearly identifiable in Satellite

building\_connection:
A) Yes, they are connected
B) No, they are separate
C) Connected by a narrow covered link only
D) Not visible in Satellite

roof\_form:
A) Flat roof
B) Gabled roof
C) Arched/curved roof
D) Mixed/unclear roof form

=========================================
OUTPUT FORMAT REQUIREMENTS

Output ONLY valid JSON (no markdown, no extra text).
Generate EXACTLY 3 objects.

Each object format:
\{
  ``question'': ``... (must include Building A/B description here when needed) ...'',
  ``options'': [``A) ...'', ``B) ...'', ``C) ...'', ``D) ...''],
  ``answer'': ``A'',
  ``category'': ``distance\_to\_camera''
\}

IMPORTANT:
- Exactly 3 objects.
- Randomize question order.
- No explanations.
\end{tcolorbox}

\subsubsection{Satellite-to-Ground (S2G) Prompt}

In S2G tasks, the solver only observes the satellite image during evaluation,
while the answer is verified from the street-view image. Again, we use two
prompt variants: one for the FOV-subset with an observation arrow, and one for
the CVUSA-subset with fixed panorama alignment.

\paragraph{S2G-FOV Prompt.}
\begin{tcolorbox}[breakable]
You are an expert dataset annotator creating Satellite-to-Street-View
cross-view reasoning multiple-choice questions.

Inputs:
1) Satellite map (top-down) with an observation point and an arrow
   (camera forward direction).
2) Street-view image captured at the same point, facing exactly the arrow
   direction.

=========================================
GEOMETRY FRAME (MANDATORY)

Forward = arrow direction.
Left/Right = viewer's left/right while facing forward.
Behind = opposite of forward.

=========================================
CORE TASK RULE (HARD)

During evaluation, the solver ONLY sees the Satellite map (with arrow).

Therefore:
Each question MUST be solvable using ONLY satellite + arrow geometry.
Street-view is used ONLY to VERIFY the correct answer after the question
is formed from satellite.

HARD RULE (Street-View Answer Determination):
- The correct answer MUST be determined by directly checking the Street-View image.
- Never guess the answer from satellite.
- If Street-View does not clearly confirm the correct option -> DISCARD.

QUALITY OVER QUANTITY:
- It is ALWAYS better to output fewer questions than low-quality questions.
- If fewer than 6 high-quality questions are possible, output fewer.

=========================================
GLOBAL HARD BANS (ABSOLUTE)

NO vehicles, cars, people, animals, temporary objects.
NO compass directions.
NO coordinates/building numbers.
NO ``unclear/undecidable'' options.

ABSOLUTE BAN: STREET-VIEW-ONLY TARGET DEFINITIONS
Targets MUST be defined using satellite-only geometry.

ABSOLUTE BAN: APPEARANCE WORDS IN TARGET DEFINITIONS
Do NOT use any appearance words to identify a target.

ABSOLUTE BAN: DIRECTION WORDS IN TARGET DEFINITIONS
Do NOT use left, right, center, middle, front, back, behind, ahead,
straight ahead in target descriptions.

ABSOLUTE BAN: ANSWER LEAKAGE
Do NOT include words that imply the answer:
large/small, tall/short, taller/higher, main/minor, occluded/blocked/hidden.

=========================================
OUTPUT

Generate exactly 3 questions total whenever possible.

Category diversity requirement (HARD):
- The 3 questions MUST come from 3 DISTINCT categories.
- Do NOT repeat a category.

If you cannot generate 3 high-quality, valid questions that meet all rules,
output fewer than 3 rather than low-quality items.

=========================================
ALLOWED CATEGORIES (ONLY THESE)

1) vegetation\_sector
2) occlusion\_binary
3) relative\_position
4) height\_comparison
5) facade\_color
6) ground\_material

=========================================
TARGET DEFINITION REQUIREMENT (CRITICAL)

Every target MUST be:
- clearly identifiable in satellite
- uniquely disambiguated using satellite-only cues such as:
  distance to camera, alignment with the arrow ray, adjacency, enclosure,
  size ranking (only if it does NOT leak the answer).

=========================================
CATEGORY RULES

1) vegetation\_sector (4 options)
Question template:
``What is the dominant vegetation type in the [sector] along the arrow direction?''
Options EXACTLY:
A) Trees
B) Shrubs / bushes
C) Grass / lawn
D) No obvious vegetation

2) occlusion\_binary (2 options ONLY)
Question template:
``Is [satellite-defined target] occluded in Street-View?''
Options EXACTLY:
A) Occluded (blocked)
B) Not occluded (clear view)

3) relative\_position (4 options)
Question template EXACTLY:
``Facing forward (the arrow direction), where does [satellite-defined target]
appear in Street-View?''
Options EXACTLY:
A) Left side
B) Right side
C) Near the center (mostly straight ahead)
D) Not visible in the forward view

4) height\_comparison (4 options, STRICT)
Question template:
``Compare the heights of (1) [Building A defined by satellite geometry] and
(2) [Building B defined by satellite geometry]. Which is taller in Street-View?''
Options EXACTLY:
A) The first building is taller
B) The second building is taller
C) They are about the same height
D) Only one of them is visible

5) facade\_color (STRICT CONDITIONAL)
Question template:
``What is the dominant facade color of [satellite-defined building] in Street-View?''
Options EXACTLY:
A) Light (white/cream/light gray)
B) Medium (tan/brown)
C) Dark (dark gray/black)
D) Red / brick-like

6) ground\_material (4 options)
Question template:
``What is the dominant ground material in the forward sector along the arrow direction?''
Options EXACTLY:
A) Asphalt
B) Concrete
C) Grass
D) Bare soil / dirt

=========================================
FINAL VALIDITY CHECK (MANDATORY)

For EACH question verify ALL:
1) Target is uniquely localizable using satellite-only cues.
2) Question is solvable from satellite + arrow only.
3) Correct answer is determined by checking Street-View.
4) Street-View clearly confirms the correct option.
If ANY check fails -> DISCARD that question.
\end{tcolorbox}

\paragraph{S2G-CVUSA Prompt.}
The CVUSA S2G prompt uses the same general logic as the FOV version, but
replaces the observation-arrow geometry with a fixed panorama mapping centered
at the satellite-image center. Specifically, the annotator is instructed to
assume that the camera is located at the center of the satellite image and
faces the top side of the satellite map. Under this convention, the panorama
center corresponds to the top of the satellite image, the panorama left/right
quarters correspond to the left/right satellite sides, and the far panorama
edges correspond to the bottom side of the satellite image. The prompt also
restricts the allowed categories to \texttt{visibility}, \texttt{vegetation},
\texttt{height}, and \texttt{location}, and requires all buildings to be
defined using 2--4 \allowbreak satellite-operational anchors rather than semantic names.
The full prompt is shown below.

\begin{tcolorbox}[breakable]
You are an expert dataset annotator creating Satellite-to-Street-View
(Sat2Ground / Sat2Pano) cross-view reasoning multiple-choice questions
for 360$^\circ$ panorama Street-View.

You will be provided with:
- A Satellite Map (top-down, fixed up/down/left/right orientation).
- A 360$^\circ$ panorama Street-View image captured at the same location.

=========================================
CAMERA \& FIXED ORIENTATION

The panorama camera location is exactly at the CENTER of the satellite image.

All reasoning assumes:
``I stand at the satellite center and face the TOP of the satellite image.''

Panorama mapping:
- Panorama CENTER -> facing TOP of satellite
- Panorama LEFT quarter -> facing LEFT of satellite
- Panorama RIGHT quarter -> facing RIGHT of satellite
- Panorama far edges -> facing BOTTOM of satellite

=========================================
CORE RULE

During generation you see BOTH images.
During evaluation the solver sees ONLY the satellite map.

Therefore:
- Every question must be solvable using satellite evidence + mapping only.
- The correct answer must be verifiable in the panorama.
- Targets must be uniquely identifiable in satellite.
- If ambiguous -> SKIP that question type.

=========================================
ABSOLUTE FORBIDDEN

- NO vehicles, people, animals.
- NO compass directions.
- NO coordinates or IDs.
- NO invented building names.
- NO facade-color-based definitions.
- NO panorama-only descriptors to define targets.
- NO ambiguous definitions.
- NO ``cannot determine''.

=========================================
BUILDING DISAMBIGUATION (CRITICAL)

Satellite images DO NOT contain semantic building labels.

You MUST NOT invent building names.

Instead, buildings must be defined using pure satellite-operational anchors.

MANDATORY BUILDING RULES
For ANY building:
- Use AT LEAST TWO independent satellite anchors.
- In dense scenes use THREE anchors.
- If still ambiguous -> SKIP.

ALLOWED SATELLITE ANCHORS
Use 2--4 of these:
1) Footprint shape (ONLY if geometrically obvious)
2) Size ranking WITHIN A DEFINED SUBSET
3) Distance-to-center ranking
4) Road adjacency
5) Enclosure
6) Structural relation (non-directional)

BUILDING LABEL FORMAT (MANDATORY)

Single building:
Target building: [FULL satellite-operational description]

Comparison:
First building: [description]
Second building: [description]

=========================================
ALLOWED QUESTION TYPES (MAX 4)

Generate 0--4 questions.
At most ONE per type.

Types:
- visibility
- vegetation
- height
- location

=========================================
TYPE 1 --- VISIBILITY (Binary)

Question format (MUST follow exactly):

``I am standing at the satellite center and facing the top of the
satellite image. Assume the fixed mapping: panorama center corresponds
to the top of the satellite map.
In the panorama, focusing on the [PANORAMA REGION], is
[SATELLITE-DEFINED TARGET] occluded?''

[PANORAMA REGION] MUST be exactly one of:
- the center region
- the left-quarter region
- the right-quarter region
- the far edge region

Options (EXACTLY):
A) Occluded
B) Not occluded

=========================================
TYPE 2 --- VEGETATION

Question:
``I am standing at the satellite center and facing the top of the satellite image.
Assume the fixed mapping: panorama center corresponds to the top of the
satellite map. In the panorama, focusing on the [PANORAMA VIEW REGION],
what is the dominant vegetation type?''

Options:
A) Trees
B) Shrubs
C) Grass
D) No visible vegetation

=========================================
TYPE 3 --- HEIGHT COMPARISON (STRUCTURED FORMAT REQUIRED)

Question:
``I am standing at the satellite center and facing the top of the
satellite image. In the panorama, what is the relative height
relationship between the following two buildings?

Building A: [FULL satellite-operational description]
Building B: [FULL satellite-operational description]''

Options (EXACTLY):
A) Building A is taller
B) Building B is taller
C) They are approximately the same height
D) Only one building is clearly visible

=========================================
TYPE 4 --- LOCATION (NO DIRECTION LEAK)

Question:
``Assume the fixed mapping: panorama center corresponds to the top of
the satellite map; panorama left-quarter corresponds to the left of
the satellite map. In which panorama region does the target building appear?''

Options:
A) The center region
B) The left-quarter region
C) The right-quarter region
D) The far edge region

=========================================
OUTPUT FORMAT

Output ONLY valid JSON:

[
  {
    ``question'': ``...'',
    ``options'': [``A) ...'', ``B) ...'', ``C) ...'', ``D) ...''],
    ``answer'': ``A'',
    ``category'': ``visibility | vegetation | height | location''
  }
]

No explanations.
Randomize order.
Maximum 4 questions.
Minimum 0.
\end{tcolorbox}

\subsubsection{Cross-view Grounding Data Generation}

In addition to question–answer tasks, CVSBench also includes a
cross-view grounding task that requires localizing the same object
across satellite and street-view images.
The grounding annotations are built on top of previously annotated
bounding boxes in both views. For each object instance, annotators
first mark the object location using a bounding box. Instead of
cropping the image region, we directly provide the full image with
the bounding box highlighted in red and ask a large multimodal model
to generate a short description of the object inside the box.
This design preserves the surrounding visual context while clearly
indicating the target object region, allowing the model to generate
more accurate and grounded descriptions.
Specifically, two prompt variants are used depending on the view
source.

\paragraph{Satellite-view description prompt.}

For satellite images, the prompt instructs the model to describe
top-down structural properties such as roof appearance, building
footprint shape, and distinctive architectural structures.

\begin{tcolorbox}[breakable]
Describe what you can see inside the red bounding box.

Focus on:
- the roof (color, material, shape)
- the building footprint and layout
- any distinctive features

Write 2–3 complete sentences with all relevant details.

Example:
"A rectangular single-story residential building with a dark gray
shingled gabled roof. The structure has a simple rectangular footprint
with a small attached carport or garage section visible on one side.
The building appears to be in good condition with clearly defined edges."

\end{tcolorbox}

\paragraph{Street-view description prompt.}

For street-view images, the prompt focuses on facade-level appearance
cues such as wall materials, windows, doors, and architectural style.
The description is intentionally concise to avoid introducing
unnecessary ambiguity.

\begin{tcolorbox}[breakable]
Please describe what you see in the red bounding box.
This is a street-level view.

Focus on:
- wall colors
- windows or doors
- architectural style
- distinctive features

Keep your description between 20–30 words.

Example:
"A single-story house with light beige walls and dark brown trim.
Features symmetrical windows and a covered front entrance."

Description:
\end{tcolorbox}

The generated descriptions serve as textual queries for the grounding
task. During evaluation, the model is given the image together with
the object description and must identify the corresponding region in
the other view.

\subsubsection{Viewpoint Localization Tasks}

In addition to cross-view VQA and grounding, CVSBench includes two
viewpoint localization tasks for the FOV-subset, where explicit camera
positions and viewing directions are manually annotated on the satellite
image. These tasks are constructed directly from the human-annotated
arrows and therefore do not require large multimodal models.

\paragraph{View-Arrow.}
This task takes as input a street-view image and a satellite image with
four candidate arrows. One arrow corresponds to the ground-truth camera
location and viewing direction of the street-view image, while the
other three serve as distractors. These distractor arrows are generated
randomly under simple geometric constraints to avoid trivial or
implausible candidates. In particular, we avoid arrows with nearly
horizontal or vertical directions and avoid placing candidate viewpoints
in obviously unsuitable regions, such as non-flat areas. The model must
select the arrow that correctly matches the given street-view image.

\paragraph{View-Image.}
This task takes as input a satellite image with a single viewpoint
arrow and four candidate street-view images. Only one street-view image
corresponds to the camera position and orientation indicated by the
arrow, while the remaining three are distractors sampled from other
views in the same FOV scene. The model is required to identify which
street-view image matches the arrow-indicated viewpoint.

\subsection{Representative Dataset Examples}

\begin{figure}[!htbp]
\centering
\includegraphics[page=1,width=\textwidth]{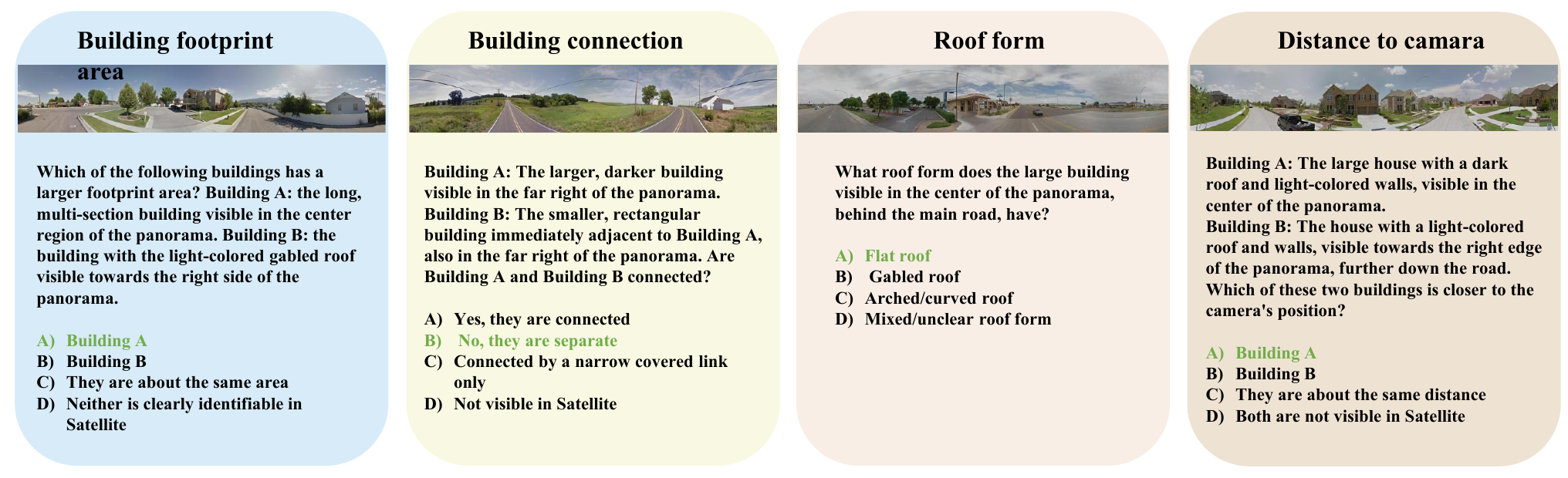}\\[4pt]
\includegraphics[page=2,width=\textwidth]{pic/app_1.pdf}
\caption{
Representative cross-view VQA examples from the CVUSA-subset, including
Ground-to-Satellite and Satellite-to-Ground question types.
}
\label{fig:dataset_examples_cvusa}
\end{figure}

\begin{figure}[!htbp]
\centering
\includegraphics[page=1,width=\textwidth]{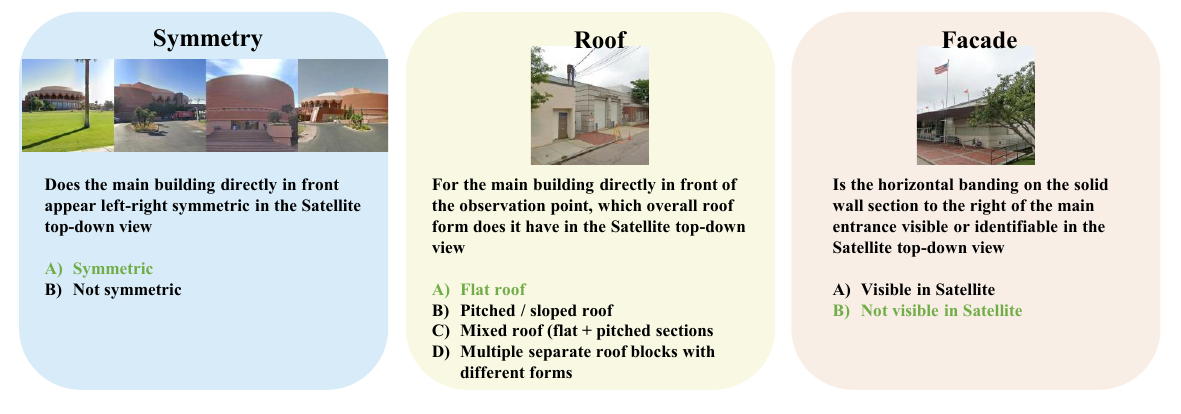}\\[4pt]
\includegraphics[page=2,width=\textwidth]{pic/app_2.pdf}\\[4pt]
\includegraphics[page=3,width=\textwidth]{pic/app_2.pdf}
\caption{
Representative cross-view VQA examples from the FOV-subset, including
facade visibility, roof form, symmetry, occlusion reasoning, relative
position, facade color, ground material, height comparison, and
vegetation reasoning.
}
\label{fig:dataset_examples_fov}
\end{figure}

\begin{figure}[!htbp]
\centering
\includegraphics[page=1,width=\textwidth]{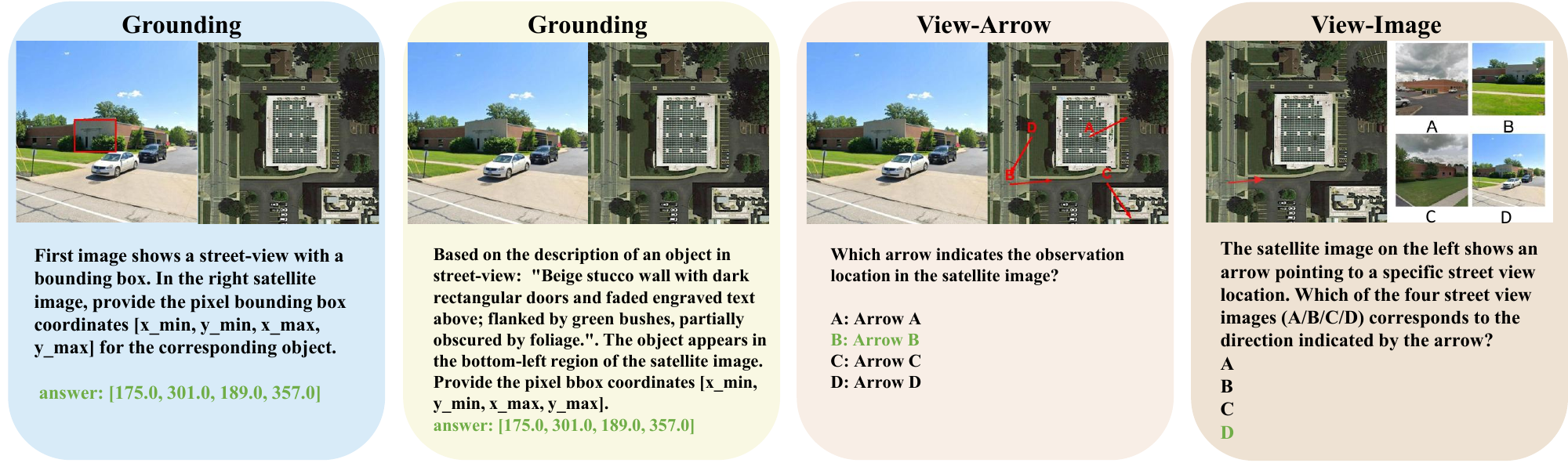}
\caption{
Representative examples of the grounding and viewpoint localization
tasks. The figure shows one representative cross-view grounding
example together with the View-Arrow and View-Image tasks. For
brevity, only one grounding example is shown, since the grounding task
format is consistent across subsets.
}
\label{fig:dataset_examples_view}
\end{figure}

Representative task examples from CVSBench are shown in
Figs.~\ref{fig:dataset_examples_cvusa}--\ref{fig:dataset_examples_view}.

\subsection{Spatial Imagination Generation with Nanobanana}

To provide additional spatial imagination cues for cross-view reasoning,
we generate auxiliary 3D miniature views using a Nanobanana-style
image generation pipeline. The generated image is designed as a clean
isometric architectural miniature that highlights structural volume,
depth ordering, and occlusion relations.

The generation pipeline supports both satellite images and street-view
images as inputs. Different prompts are used depending on the source
view. In all cases, the output is a single square isometric rendering
with a fixed resolution of 1024$\times$1024 pixels.

\paragraph{Satellite imagination prompt.}

For satellite inputs, the generator reconstructs a high-fidelity
3D isometric miniature from the top-down map layout while preserving
structural geometry and spatial relationships.

The exact prompt used in our pipeline is shown below.

\begin{tcolorbox}[breakable]
A high-fidelity 3D isometric miniature model reconstructed from a
satellite map layout, blending accurate spatial volume with a high-end
toy aesthetic.

Structural \& Spatial Logic:
- Reconstruct architectural structures with distinct vertical elevations.
- Emphasize structural occlusion and depth perception to clearly
  distinguish between tall buildings, low-level ground, and sunken roads.
- Ensure sharp edge definition to separate foreground objects from the
  background.

Composition \& Base:
- Place the scene on a thick, clean, rounded-corner isometric tile
  (white or light grey) acting as a minimalist pedestal.
- Use a strict 45-degree top-down bird's-eye perspective.
- The final image must have a 1:1 aspect ratio.

Material \& Lighting:
- Use refined PBR materials with realistic textures such as concrete,
  asphalt, and metal roofs.
- Lighting should balance soft global illumination with directional
  sunlight to create clear shadows that highlight 3D volume.

Visual Style:
- Clean, fresh, and tangible.
- The final result should resemble a premium architectural miniature
  or clay render.
\end{tcolorbox}

\paragraph{Street-view imagination prompt (single-view).}

For most Ground-to-Satellite (G2S) question types, imagination views
are generated from a single street-view image. The generator converts
the street-level observation into a consistent 3D isometric miniature
while preserving the spatial relationship between foreground and
background structures.

The prompt used for single-view street imagination is shown below.

\begin{tcolorbox}[breakable]
A 3D isometric miniature rendering based on multiple street-level perspectives, emphasizing structural occlusion and depth perception. The model accurately integrates architectural features from different angles onto a centered isometric tile. Focus on the overlapping relationship between buildings: foreground objects must be clearly separated from background structures through shadow depth and edge definition. 45-degree bird's-eye view, 1:1 ratio. Textures are refined PBR materials. The lighting is designed to highlight 3D volumes, ensuring that the relative positions and 'front-to-back' distances are visually intuitive for information analysis. Clean, fresh, high-end toy aesthetic.
\end{tcolorbox}

\paragraph{Multi-view imagination for symmetry questions.}

Certain Ground-to-Satellite questions require reasoning about building
symmetry. A single street-view image may not provide sufficient facade
coverage to reliably infer symmetric structures. Therefore, for
symmetry-related questions we allow multiple street-view inputs.

Specifically, up to four street-view images captured at the same
location are jointly provided to the
generator. These images are integrated to reconstruct a coherent 3D
miniature that better reflects the global building layout and
potential symmetry.

The same prompt as the single-view street setting is used, but the
generator receives multiple images simultaneously and merges their
structural cues into a unified isometric scene.

\begin{figure}[!htbp]
\centering
\includegraphics[page=1,trim=0cm 2.3cm 0cm 0.6cm,clip,width=0.95\textwidth]{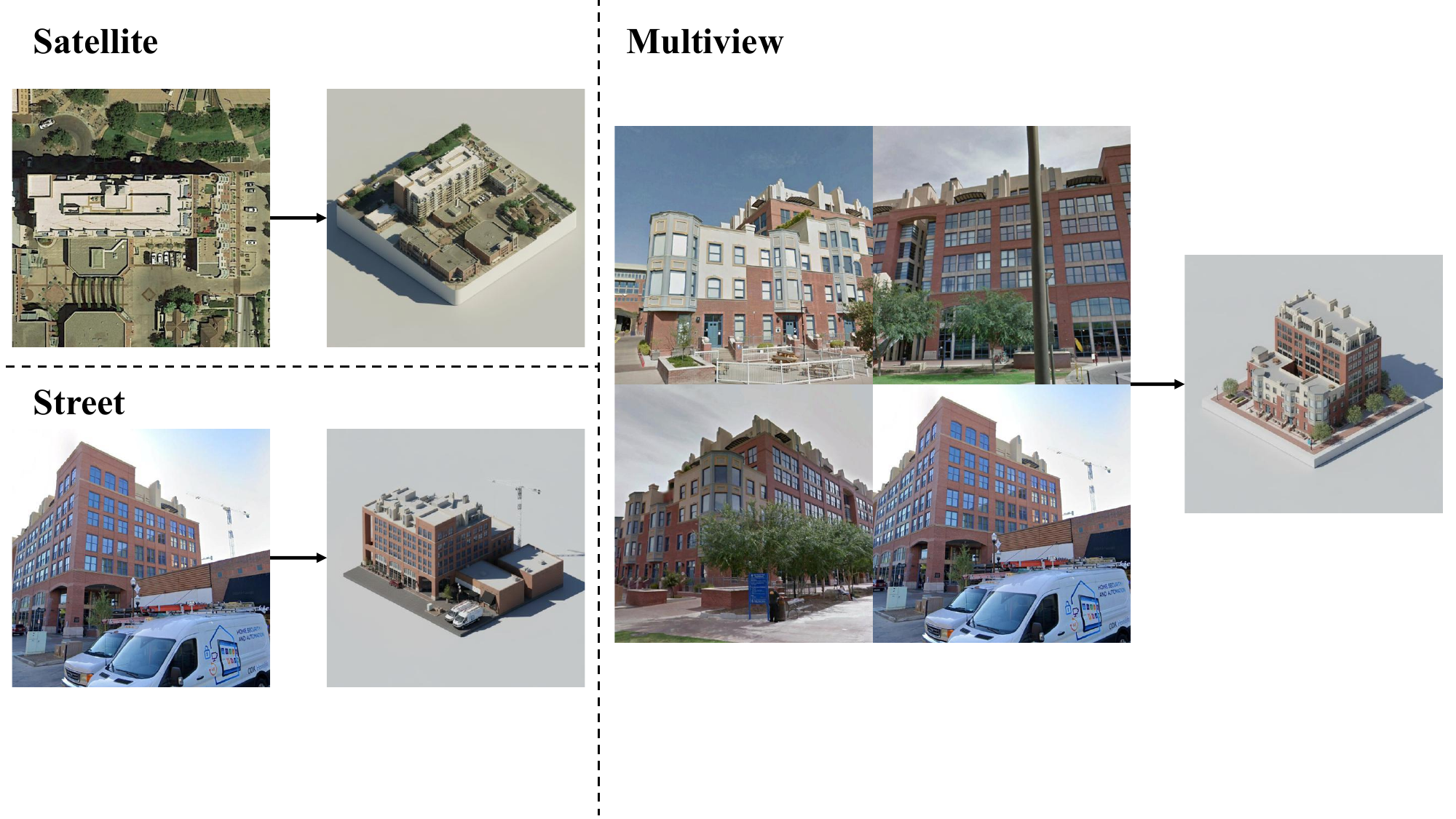}
\caption{
Illustration of the three spatial imagination settings used in our
Nanobanana-style generation pipeline. Left top: satellite-view input
and the corresponding generated isometric miniature. Left bottom:
single street-view input and the corresponding generated miniature.
Right: multi-view street-view inputs used for symmetry-related
questions and the resulting unified isometric miniature. The generated
miniatures are used as auxiliary visual cues for cross-view reasoning
and are not treated as ground-truth annotations.
}
\label{fig:nanobanana_examples}
\end{figure}

The generated imagination views are used as auxiliary visual inputs in
cross-view reasoning experiments and are not treated as ground-truth
annotations.
\begin{tcolorbox}[breakable]
Generate an image of a single, cohesive 3D isometric miniature model by synthesizing information from the provided sequence of street-view images.

CRITICAL PERSPECTIVE INSTRUCTION:
- Viewpoint: STRICTLY 45-degree top-down bird's-eye view (Orthographic Projection).
- Constraint: DO NOT render from street level or ground level. The camera must be high above the ground.
- Composition: The building block must appear as a "floating island" or a distinct "architectural asset" on a centered isometric tile.

Integration \& Spatial Logic:
- Mentally stitch the building facades and street details from all inputs to form a complete comprehensive block understanding.
- Focus on "structural occlusion" and "depth perception": accurately resolve the overlapping relationship between buildings.
- Prioritize structural continuity when visual data conflicts.

Visual Output:
- 45-degree top-down bird's-eye perspective, 1:1 aspect ratio.
- Emphasize depth: Foreground objects (trees/poles) must be clearly separated from background structures through shadow depth and edge definition.
- Lighting: Designed to highlight 3D volumes, ensuring relative positions and 'front-to-back' distances are visually intuitive.
- Style: Clean, fresh, "high-end toy aesthetic" mixed with an "architectural clay render" style. Refined PBR textures.

Output Requirement:
- Provide ONLY the generated image. Do not output any text descriptions, explanations, or JSON.
\end{tcolorbox}
Representative examples of the three imagination settings are shown in
Fig.~\ref{fig:nanobanana_examples}.

\subsection{Quantitative Analysis of Cross-view Differences}

\begin{figure}[htbp]
    \centering
    \begin{subfigure}[b]{0.24\textwidth}
        \centering
        \includegraphics[width=\textwidth]{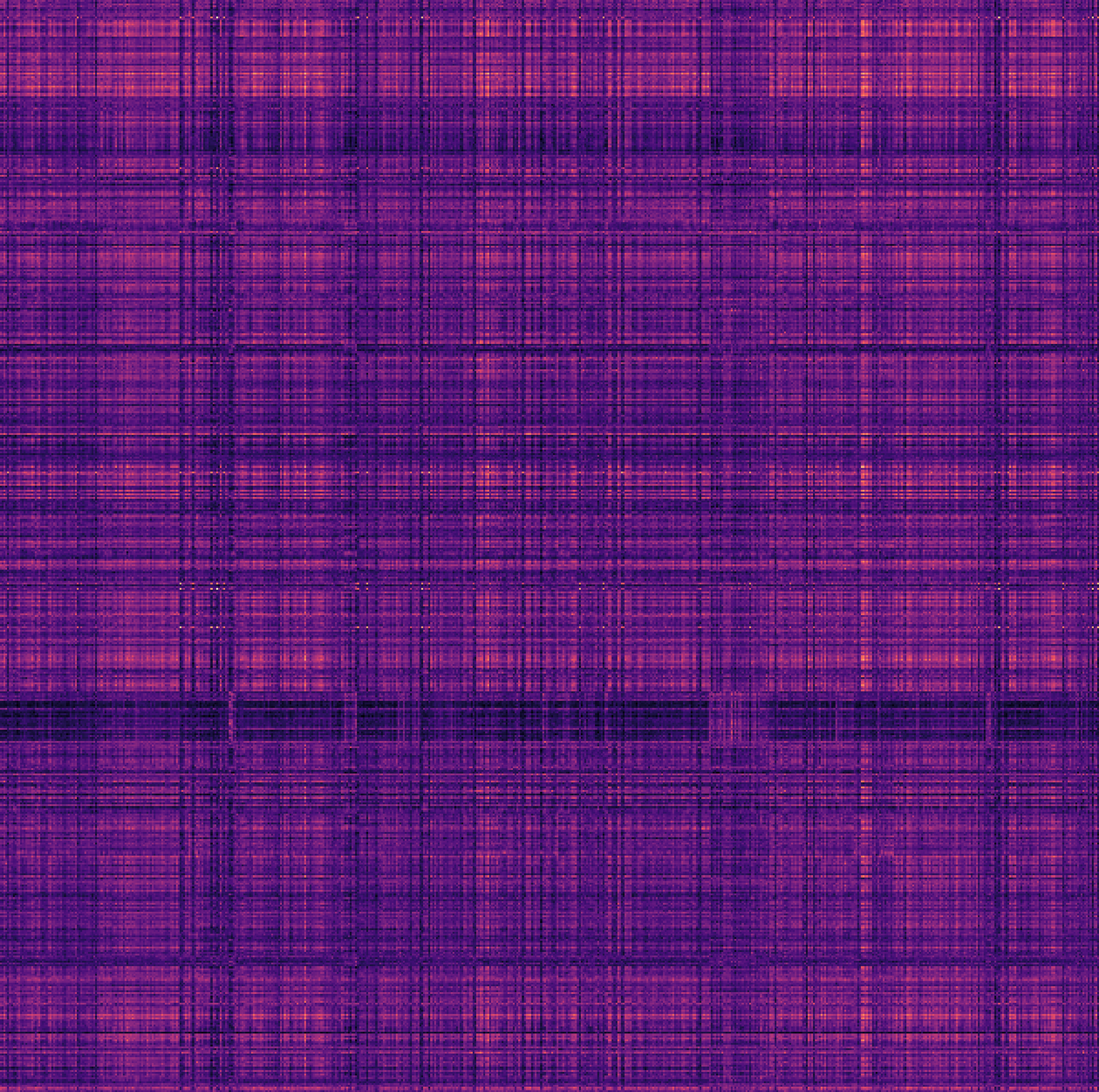}
        \caption{FOV(Ours)}
        \label{fig:heat_fov}
    \end{subfigure}
    \hfill
    \begin{subfigure}[b]{0.24\textwidth}
        \centering
        \includegraphics[width=\textwidth]{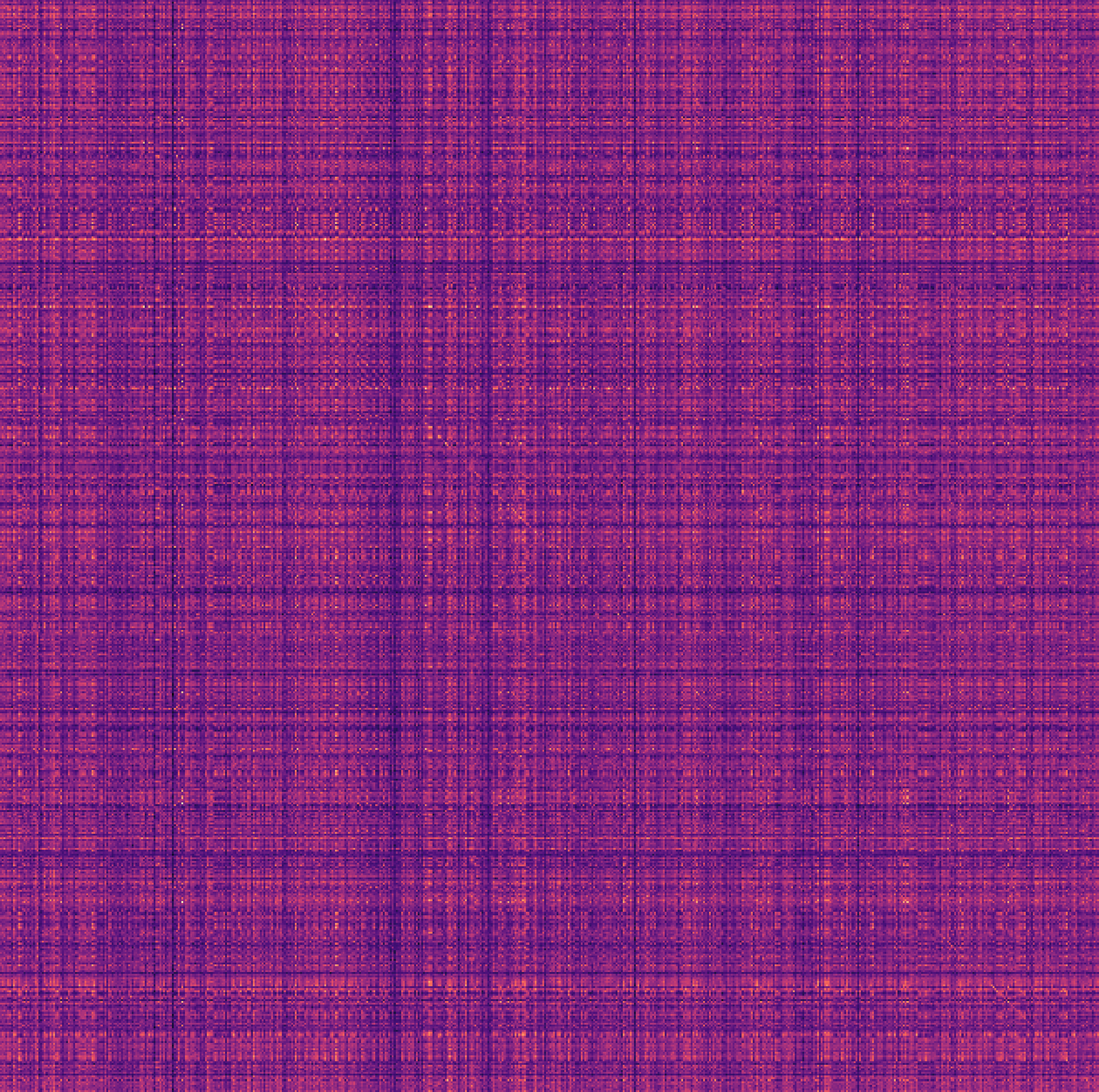}
        \caption{CVUSA(Ours)}
        \label{fig:heat_cvusa}
    \end{subfigure}
    \hfill
    \begin{subfigure}[b]{0.24\textwidth}
        \centering
        \includegraphics[width=\textwidth]{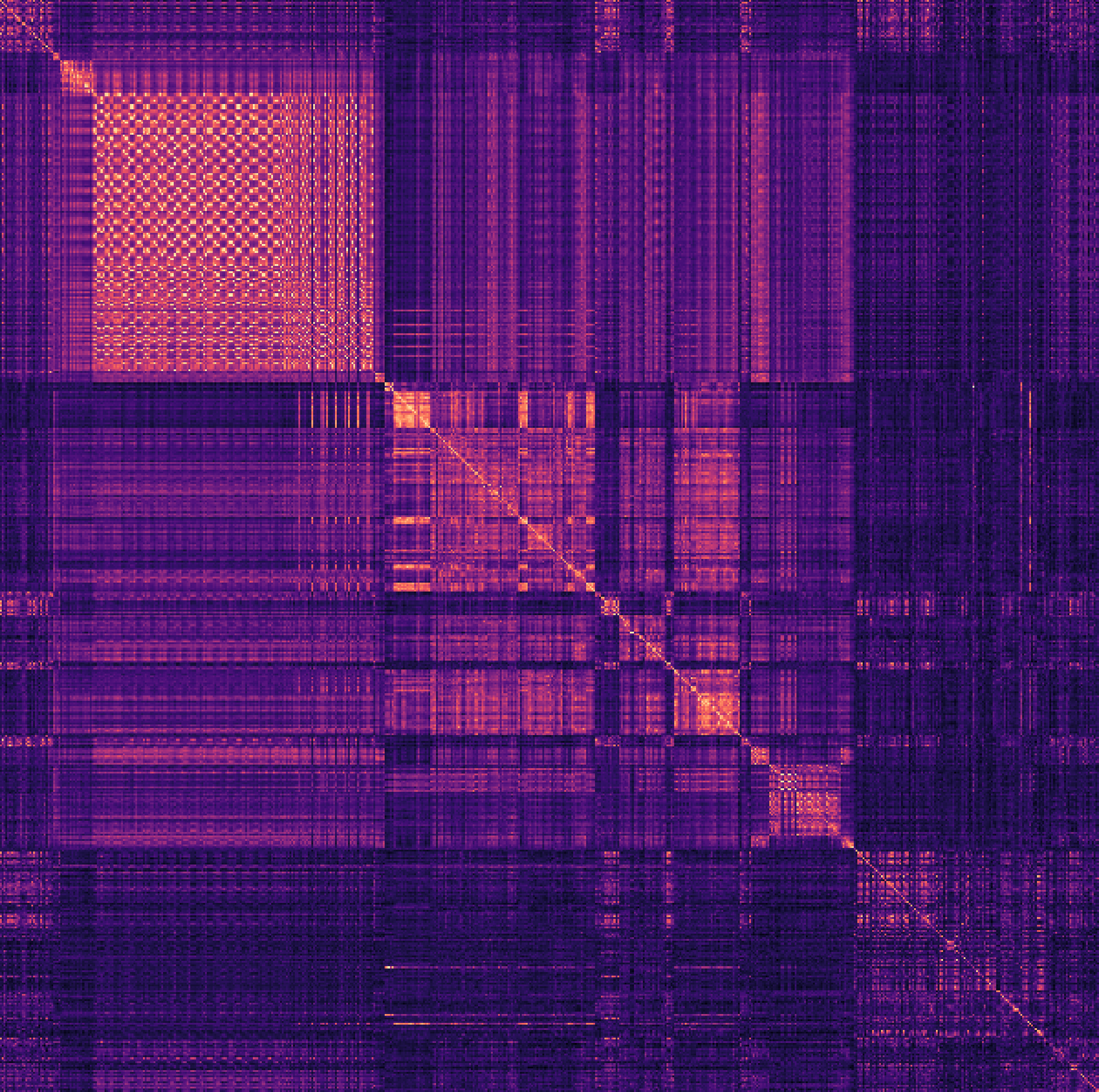}
        \caption{MindCube}
        \label{fig:heat_mindcube}
    \end{subfigure}
    \hfill
    \begin{subfigure}[b]{0.24\textwidth}
        \centering
        \includegraphics[width=\textwidth]{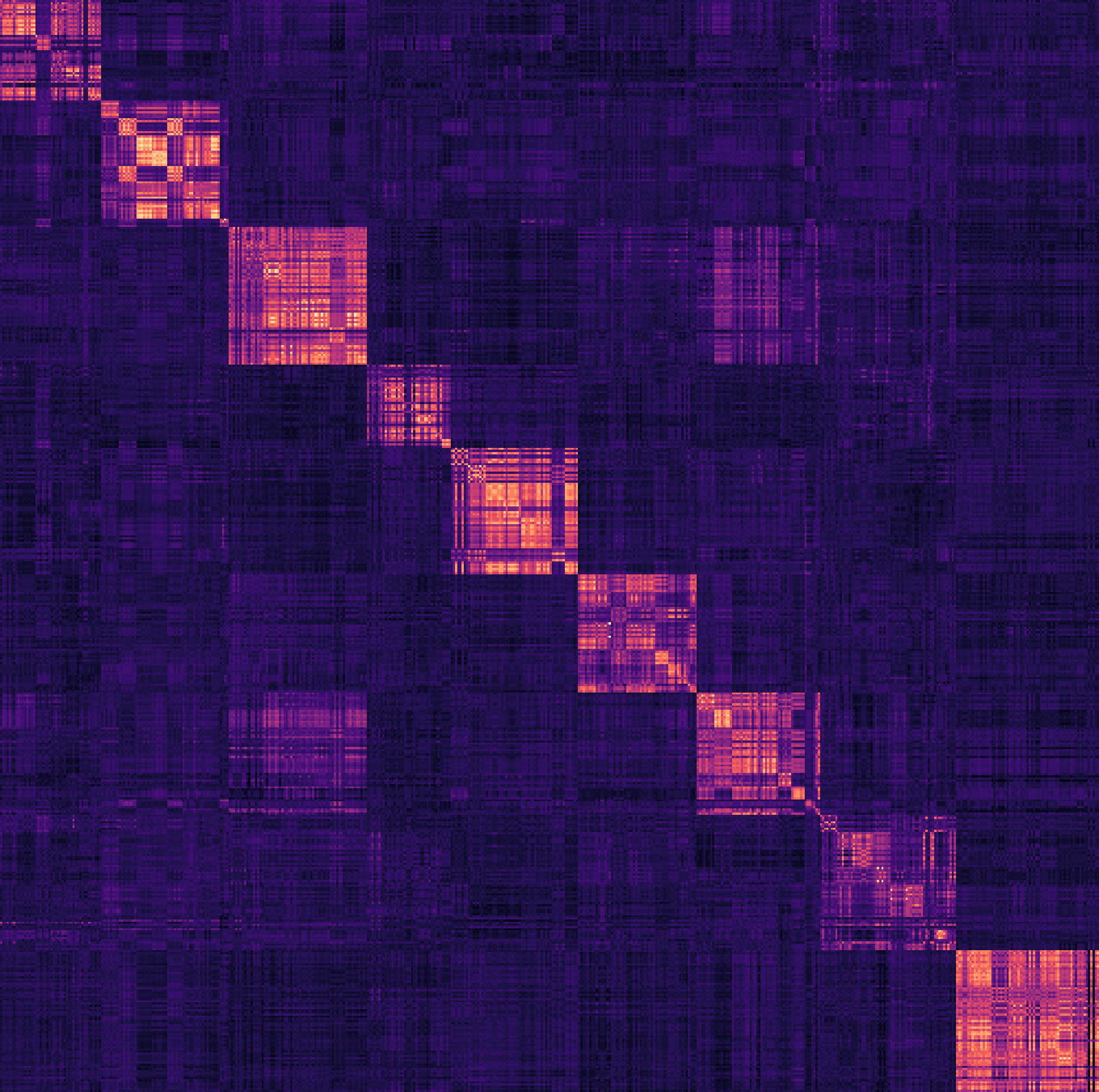}
        \caption{NLVR2}
        \label{fig:heat_nlvr2}
    \end{subfigure}

    \caption{comparison of similarity heatmaps across different datasets. Brightness indicates cosine similarity between image features extracted by CLIP. Compared with existing spatial reasoning benchmarks such as MindCube and NLVR2, our datasets (FOV and CVUSA) exhibit significantly lower off-diagonal similarity, indicating larger cross-view discrepancies and stronger viewpoint variations.}
    \label{fig:similarity_comparison}
\end{figure}

To quantitatively analyze the diversity and the extent of viewpoint variations within different datasets, we compute cosine similarity matrices based on CLIP feature representations and visualize them as heatmaps. As shown in Fig.~\ref{fig:similarity_comparison}, these heatmaps reveal the internal correlation structures of different datasets. Each row and column correspond to images in the dataset, and each entry represents the cosine similarity between the corresponding image features.

The MindCube dataset exhibits clear block-wise regions with high similarity, along with a pronounced diagonal structure. This indicates that many samples share similar visual layouts or object configurations, suggesting a relatively limited diversity of visual patterns. Such clustered structures imply that the dataset contains a considerable degree of redundancy among samples.

A similar phenomenon can be observed in NLVR2. The heatmap shows strong diagonal patterns and several regions of elevated similarity, indicating that many image pairs share comparable visual structures. This observation aligns with the construction of NLVR2, where paired images are generated under controlled conditions with relatively small visual variations.

In contrast, the FOV and CVUSA subsets of CVSBench demonstrate substantially lower off-diagonal similarity. The heatmaps appear more uniformly distributed with fewer concentrated bright regions, indicating larger feature discrepancies across samples. This pattern suggests that images within our datasets exhibit significantly greater viewpoint differences, posing a more challenging setting for spatial reasoning under large cross-view transformations. Such characteristics better reflect real-world scenarios, where observations of the same location may originate from drastically different viewpoints.
\subsection{Ethics, Licensing, and Dataset Release}

CVSBench is constructed using images from two publicly available
cross-view datasets: CVUSA and University-1652.

The CVUSA dataset provides large-scale pairs of ground-view and aerial
images collected across the United States and is publicly accessible
through online repositories. The University-1652 dataset is distributed
for academic research use and can be obtained by submitting a request
to the dataset authors.
CVSBench does not introduce new raw imagery. All images used in our
benchmark originate from these datasets and therefore inherit their
original licensing terms.
To comply with the licensing policies of the source datasets, we do
not redistribute the raw images. Instead, we release only the benchmark
annotations required to reproduce the tasks in CVSBench. These include
viewpoint arrows, cross-view bounding boxes, grounding descriptions,
and question--answer pairs.
Researchers can obtain the original images from the official sources of
the CVUSA and University-1652 datasets and combine them with our
released annotations to reconstruct the full CVSBench benchmark.

The CVSBench annotations, evaluation scripts, and data construction
tools will be publicly released for research purposes. The dataset is
intended strictly for academic research and benchmarking of vision–
language spatial reasoning models. Any commercial use must comply with
the licensing terms of the original datasets.
The dataset mainly contains urban and suburban outdoor scenes inherited
from the source datasets and does not intentionally include sensitive
personal information. Users of the dataset are responsible for complying
with the ethical and legal requirements of the original data sources.
\section{Training Details}

\subsection{Training Environment and Hardware}

All experiments are conducted using the \texttt{verl} reinforcement
learning framework on a single node equipped with eight NVIDIA A100
GPUs. The base model is Qwen3-VL-4B, and the reinforcement learning stage
is initialized from the SFT checkpoint.

Training is implemented through the \texttt{verl.trainer.main\_ppo}
entrypoint with \texttt{algorithm.adv\_estimator=grpo}. The rollout
generation engine uses \texttt{vllm}, which allows efficient batched
generation during policy optimization.

The supervised fine-tuning stage takes approximately 30 minutes,
while the GRPO reinforcement learning stage requires about 4.5 hour
on 8$\times$A100 GPUs. Overall, the complete training pipeline
finishes within about 5 hours.

During inference, evaluating the full benchmark requires
approximately 30--40 minutes depending on the model size.

\subsection{Training Instructions}

The Chain-of-Thought (CoT) reasoning traces used for supervised
fine-tuning are automatically generated using a large multimodal model
under structured prompt templates. These prompts are designed to
simulate human cross-view reasoning under a single-view constraint,
namely, the model must reason only from the visible input image and
must not use the unseen view as reasoning evidence.

We use four prompt variants corresponding to the four task settings in
CVSBench: Ground-to-Satellite (CVUSA), Ground-to-Satellite (FOV),
Satellite-to-Ground (CVUSA), and Satellite-to-Ground (FOV). Although
the visible input views differ across these tasks, all prompts follow a
similar reasoning structure: (1) concrete observations from the visible
image, (2) identifying the required cross-view mapping, (3) projecting
the observed cues into a plausible cross-view spatial representation,
and (4) performing the final reasoning before producing the answer.

\noindent\textbf{Example prompts for CoT data generation.}
The two CoT variants are generated under different input settings.
For imagination-based CoT generation, both the question-side input
image and its corresponding complementary view are provided during data
generation. However, the prompt explicitly instructs the model to
reason only from the image that would be visible at test time, while
the complementary view is used only for silent verification of the
final answer.

For structured CoT generation, only the single-view image used in the
question (either street-view or satellite depending on the task) is
provided. The reasoning process must therefore rely solely on the
visible evidence in that image.
\noindent\textit{Imagination-based CoT prompt.}

\begin{tcolorbox}[breakable]
\small

You are generating supervised Chain-of-Thought (CoT) training data.

Input (two images provided simultaneously):

Image 1: Street-view image (ground-level)

Image 2: Satellite image (top-down)

Your task:
Simulate a human reasoning process where the person can ONLY see the
Street-view image during reasoning.

You must:

Start from the Street-view image.

Observe concrete visual details step by step.

Infer how the target structure/object would likely appear in a
Satellite top-down view.

Then output the final answer.

IMPORTANT SIMULATION RULE:
During reasoning, assume you do NOT have access to the Satellite image.
You may look at the Satellite image only silently to verify the final
answer, but you must NOT mention any satellite-visible details.

Hard Evidence Constraint (the only strict rule)

ALL reasoning evidence must come strictly from the Street-view image.

Do NOT cite or describe any satellite-visible details.

Do NOT guess specific brand names or place identities. Describe only
visible structural features.

If a detail is unclear from the Street-view image, explicitly mention
uncertainty instead of inventing information.

Do NOT output any option letter or final conclusion.

Do NOT use decisive language such as ``therefore the answer is''.

The entire output (reasoning + final answer) must be within 200 words.

Output format (MANDATORY)

Step 1: Street-view observations -- Describe key visible spatial cues
from the Street-view image.

Step 2: Cross-view mapping intention -- In one sentence, explain what
type of top-down inference this question requires.

Step 3: Projection / imagination -- Explain how the observed
Street-view cues could translate into a plausible Satellite top-down
representation. Mention possible ambiguity sources.

Step 4: Reasoning process -- Provide logical reasoning based on the
above steps. Do NOT reveal the final answer here.

\texttt{<answer> Output ONLY the final option letter: A or B or C or D </answer>}

\end{tcolorbox}

\noindent\textit{Structured CoT prompt.}

\begin{tcolorbox}[breakable]
\small

You are generating structured cross-view reasoning data for CVUSA G2S.

Use ONLY the street-view image for reasoning.

The entire output must be within 200 words.

Allowed question types:

Distance to camera (map→view mapping)

Footprint area comparison (not vegetation vs vegetation)

Building connection

Roof shape

Forbidden:

Any road connection question

Any non-roof generic shape question

Vegetation-to-vegetation area comparison

Output structure:

Overview

Briefly describe scene elements relevant to the question.

Key Objects

Provide 1–3 critical objects:

label

bbox: [x1,y1,x2,y2] (pixel coordinates)

description (only visible evidence related to distance / footprint / roof / connection)

Spatial Reasoning

Infer how the object would project into top-down view (area, adjacency,
roof geometry, perspective depth cues). Mention uncertainty if
perspective distortion exists.

\end{tcolorbox}

During supervised fine-tuning and reinforcement learning, we use two
instruction styles to train the model for cross-view spatial reasoning.
The first style is an imagination-based chain-of-thought (CoT)
instruction, which encourages the model to mentally project the visible
scene into the corresponding cross-view representation. The second style
is a structured reasoning instruction, which organizes the output into
explicit semantic sections. In both settings, the model is constrained
to reason only from the visible evidence in the given image, avoid
hallucinating unseen details, and end with a final answer in the format
\texttt{<answer> A/B/C/D </answer>}.

\noindent\textbf{Training instruction templates.}

\noindent\textit{Imagination-based instruction.}

\begin{tcolorbox}[breakable]
\small

You are a vision-language model for spatial reasoning. Input: one image
(either Street-view OR Satellite) and one multiple-choice question
(A/B/C/D).

Core rule (STRICT): all reasoning evidence must come ONLY from the given
image; do NOT reference or assume any information from an unseen view.

Evidence constraints: describe only concrete, visible cues in the image;
do NOT hallucinate; do NOT guess place identities, store names, or
brand names; if a detail is unclear, briefly state uncertainty but still
choose the best option.

Output constraints: in reasoning, do NOT mention option letters
(A/B/C/D) or reveal the final choice; total output must be within
200 English words; do NOT output JSON, code blocks, tool wrappers, or
system messages.

Mandatory output format (must match exactly):

Step 1: Observations \\
Step 2: Cross-view mapping intention \\
Step 3: Projection / imagination \\
Step 4: Final reasoning \\
\texttt{<answer> A/B/C/D </answer>}

WARNING: You must output ONLY in the specified format. Any deviation
from the format will be considered invalid.

\end{tcolorbox}

\noindent\textit{Structured reasoning instruction.}

\begin{tcolorbox}[breakable]
\small

You are a professional cross-view image understanding model. You must
perform structured reasoning and strictly follow these rules.

General Constraints: Output must strictly follow the three-section
structure: Overview $\rightarrow$ Key Objects $\rightarrow$ Spatial
Reasoning. Keep the entire output within 200 words. Use only visible
visual evidence from the image; do not guess objects not present. Use
objective and concise language, describing only observable features.

Core Task:

Overview: Briefly describe key scene elements relevant to the question.

Key Objects: List 1--3 critical objects, including label, bounding box
(bbox), and visual descriptions relevant to the question
(e.g., vegetation type, material, roof shape, distance, etc.).

Spatial Reasoning: Infer spatial relationships based on visible
evidence, explain perspective mapping (e.g., street-view to satellite
top-down projection), and mention uncertainty caused by occlusion or
perspective distortion.

Finally output: \texttt{<answer> A/B/C/D </answer>}

WARNING: You must output ONLY in the specified format. Any deviation
from the format will be considered invalid.

\end{tcolorbox}

The other task variants follow the same reasoning structure but impose
different constraints depending on the task setting, such as visibility
prediction, vegetation type inference, façade features, or direction
reasoning.





\subsection{Training Data Split}

The training dataset is evenly divided into two disjoint subsets.
One half is used for supervised fine-tuning (SFT), where the model is
trained with Chain-of-Thought (CoT) supervision under the instruction
templates described above. The other half is used for reinforcement
learning (RL) with GRPO. This separation avoids data leakage between
the two training stages and enables a cleaner analysis of the effect of
RL optimization beyond supervised initialization.

\subsection{GRPO Training Setup}

After supervised initialization, the model is further optimized using
Group Relative Policy Optimization (GRPO). During each training step,
multiple candidate responses are generated for every prompt and used to
estimate relative advantages for policy updates.

Response generation for rollouts is performed using the \texttt{vllm}
engine. For each prompt, the model samples four candidate responses
($G=4$). These responses are used to compute policy gradients.

KL regularization is applied to stabilize policy updates. The KL
coefficient is set to 0.01 and the low-variance KL formulation is used.
The KL term is applied within the policy optimization objective but is
not included directly in the reward function.

The actor network is optimized with a learning rate of
$1\times10^{-6}$. The training batch size is 64 samples. GRPO optimization
uses a mini-batch size of 16 and a micro-batch size of 1 per GPU.


\subsection{Key Training Hyperparameters}

Table~\ref{tab:grpo_hparams} summarizes the key hyperparameters used
during GRPO training. These parameters correspond to the launch
configuration in the \texttt{verl} training framework.

\begin{table}[h]
\centering
\caption{Key hyperparameters used during GRPO training.}
\label{tab:grpo_hparams}
\begin{tabular}{ll}
\toprule
Parameter & Value \\
\midrule
Framework & verl \\
Algorithm & GRPO \\
Base model & Qwen3-VL-4B-SFT \\
GPUs & 8 × NVIDIA A800 \\
Nodes & 1 \\
Train batch size & 64 \\
Learning rate & $1\times10^{-6}$ \\
PPO mini-batch size & 16 \\
Micro-batch per GPU & 1 \\
KL coefficient & 0.01 \\
Responses per prompt ($G$) & 4 \\
Max prompt length & 4096 \\
Max response length & 1024 \\
Total epochs & 3 \\
Rollout engine & vllm \\
Reward function & $0.9R_{\text{acc}} + 0.1R_{\text{format}}$ \\
\bottomrule
\end{tabular}
\end{table}

\subsection{Reward Function}

The reinforcement learning reward is computed using a lightweight
task-specific scoring function implemented in \texttt{cvs\_reward.py}.
The reward consists of two components: an accuracy reward and a format
reward.

The accuracy reward checks whether the predicted answer extracted from
the \texttt{<answer>} tag matches the ground-truth answer. The format
reward checks whether the output follows the required answer format.

The final reward is computed as

\[
R = (1-\lambda)R_{acc} + \lambda R_{format}
\]

where $\lambda = 0.1$. This formulation ensures that answer correctness
remains the dominant optimization signal while encouraging valid output
formats.

\subsection{Inference Setup and Training Convergence}

To analyze the training behavior of GRPO optimization, we track the
average reward during reinforcement learning for the two reasoning
variants, \textit{Structure CoT} and \textit{Imagination CoT}. During
evaluation, inference is performed using the \texttt{vllm} engine with
an OpenAI-compatible API. The decoding temperature is set to 0.01 and
the maximum generation length is limited to 512 tokens.

\begin{figure}[htbp]
    \centering
    \includegraphics[width=0.9\textwidth]{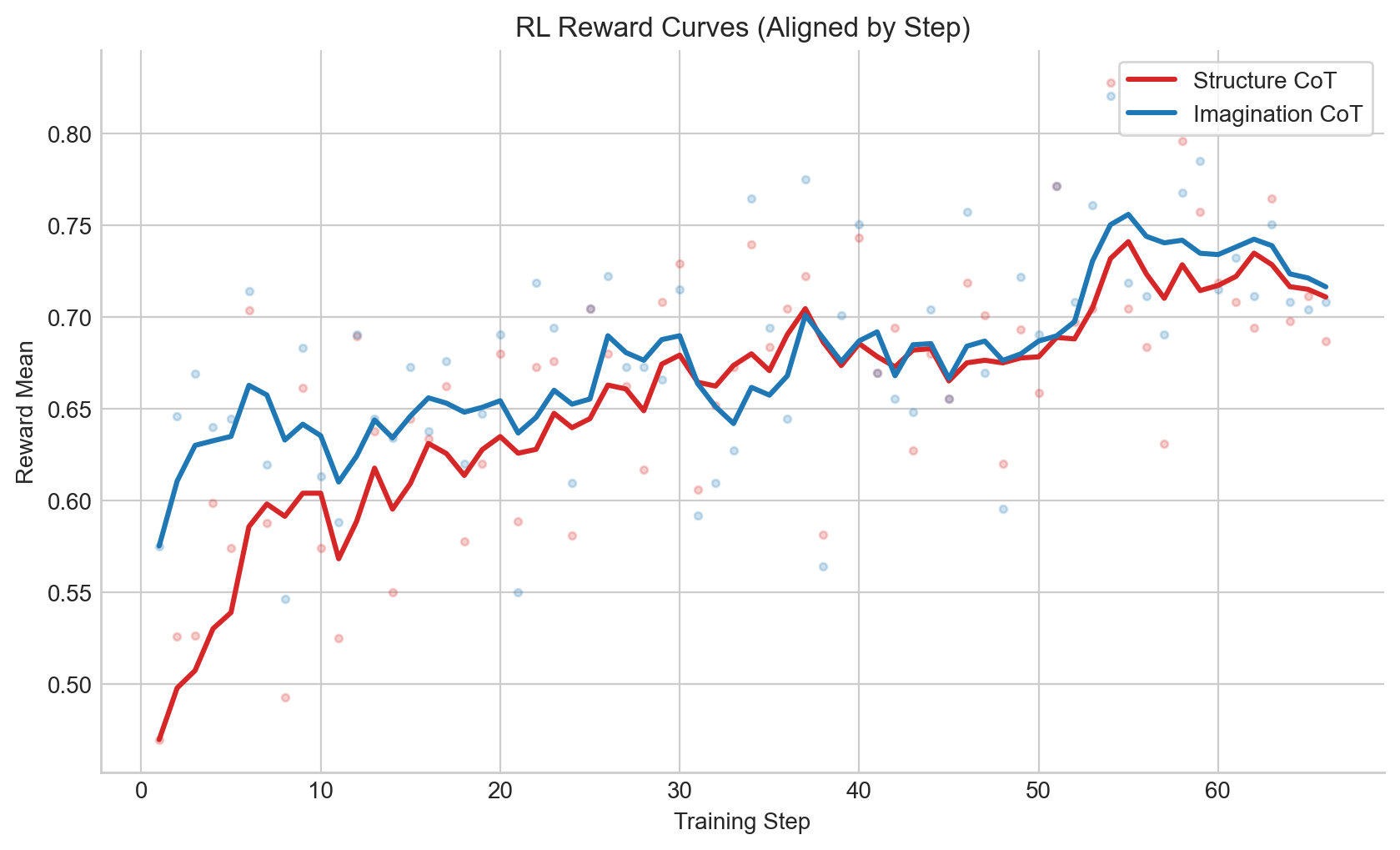}
    \caption{
    Reward curves during GRPO training for the two reasoning variants.
    Both models show steady reward improvement, while
    \textit{Imagination CoT} maintains a slightly higher reward in the
    early and middle stages. The curves converge in later training
    steps, indicating stable optimization.
    }
    \label{fig:rl_reward_curve}
\end{figure}

Figure~\ref{fig:rl_reward_curve} shows the reward curves aligned by
training step. Both variants exhibit a consistent upward trend during
training, indicating that reinforcement learning provides a stable
optimization signal. The \textit{Imagination CoT} variant starts from a
higher reward level in the early stage, suggesting stronger initial
reasoning quality under the current reward formulation. In contrast,
\textit{Structure CoT} begins with a lower reward but improves steadily
throughout training.

As training progresses, the gap between the two curves gradually
narrows and both variants converge to similar reward levels. In the
later stage, the curves stabilize with only minor fluctuations and no
evidence of reward collapse, indicating stable convergence of GRPO
optimization.

\section{Inference Details and Additional Results}

\subsection{Inference Setup and Instructions}

We evaluate five model variants, including the base model
\texttt{Qwen3-VL-4B}, two supervised fine-tuned models
(\texttt{SFT1} and \texttt{SFT2}), and two reinforcement learning models
(\texttt{RL1} and \texttt{RL2}).

Inference is conducted using the \texttt{vllm} engine through an
OpenAI-compatible local API server. Each evaluation sample consists of a
single image together with a multiple-choice question and its answer
options. The corresponding instruction template is prepended to the
input prompt during inference.
For multiple-choice evaluation, the decoding temperature is set to 0.01
and the maximum generation length is limited to 512 tokens. 

\subsection{Additional Qualitative Results}

\begin{figure}[!htbp]
\centering
\includegraphics[page=1,width=\textwidth]{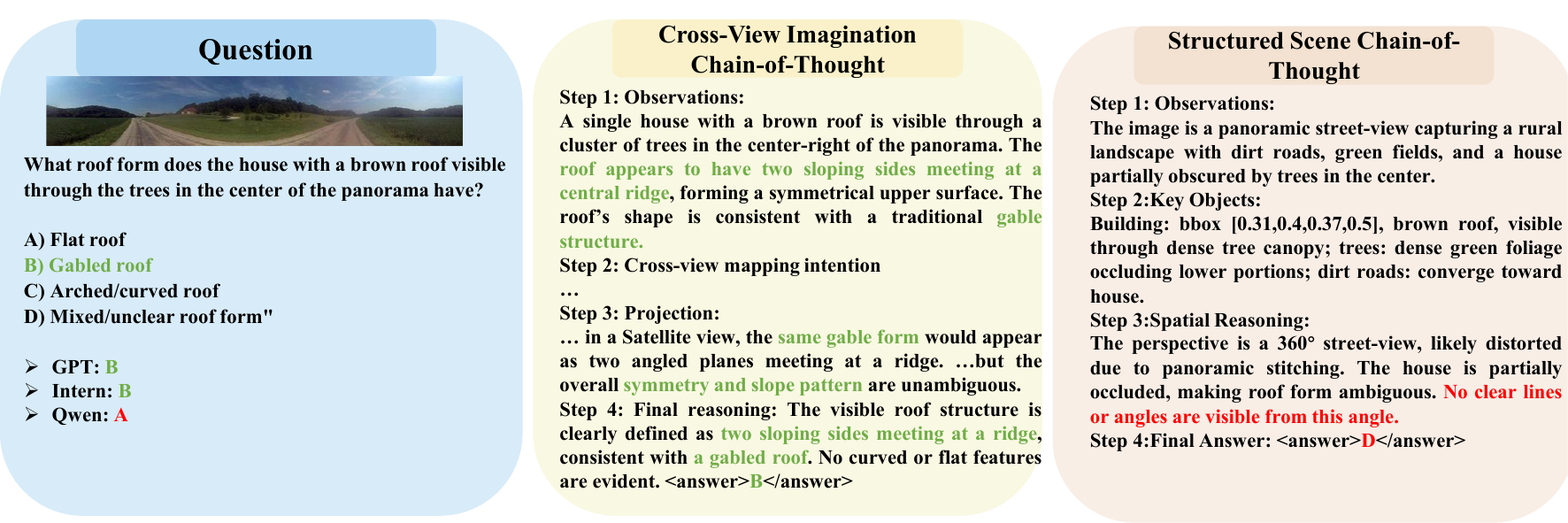}\\[4pt]
\includegraphics[page=2,width=\textwidth]{pic/app_6.pdf}\\[4pt]
\includegraphics[page=3,width=\textwidth]{pic/app_6.pdf}
\caption{
Additional qualitative comparisons between Imagination-CoT
and Structured-CoT on representative CVSBench examples
}
\label{fig:additional_qualitative_results_1}
\end{figure}

\begin{figure}[!htbp]
\centering
\includegraphics[page=4,width=\textwidth]{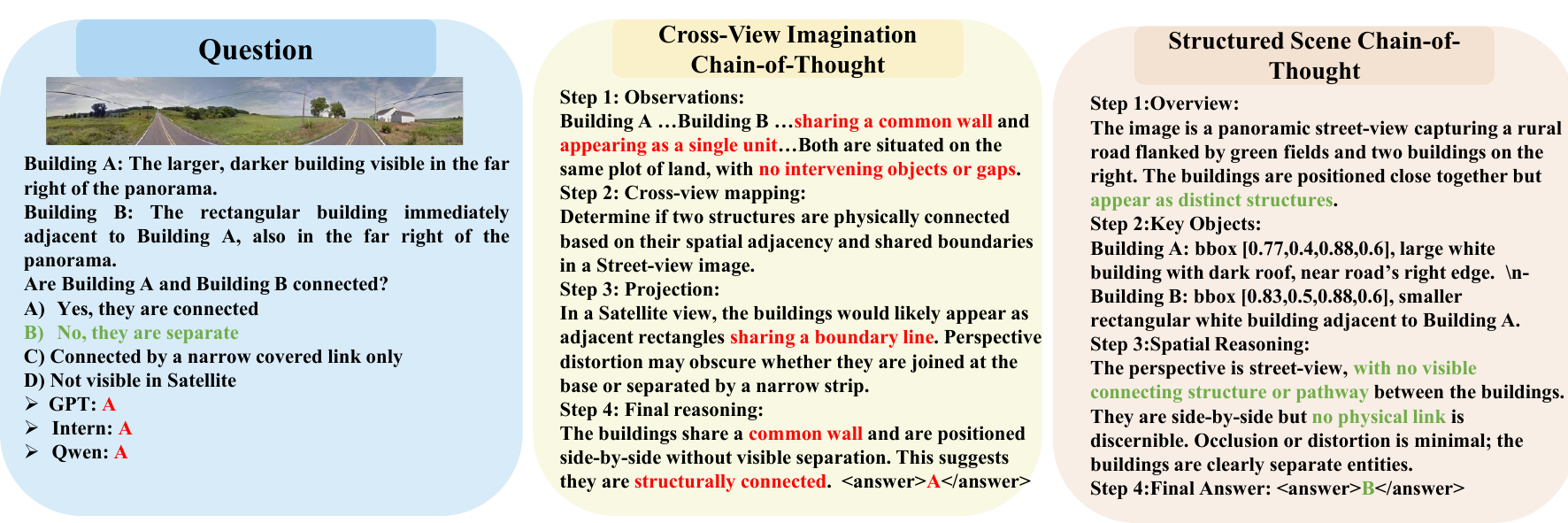}\\[4pt]
\includegraphics[page=5,width=\textwidth]{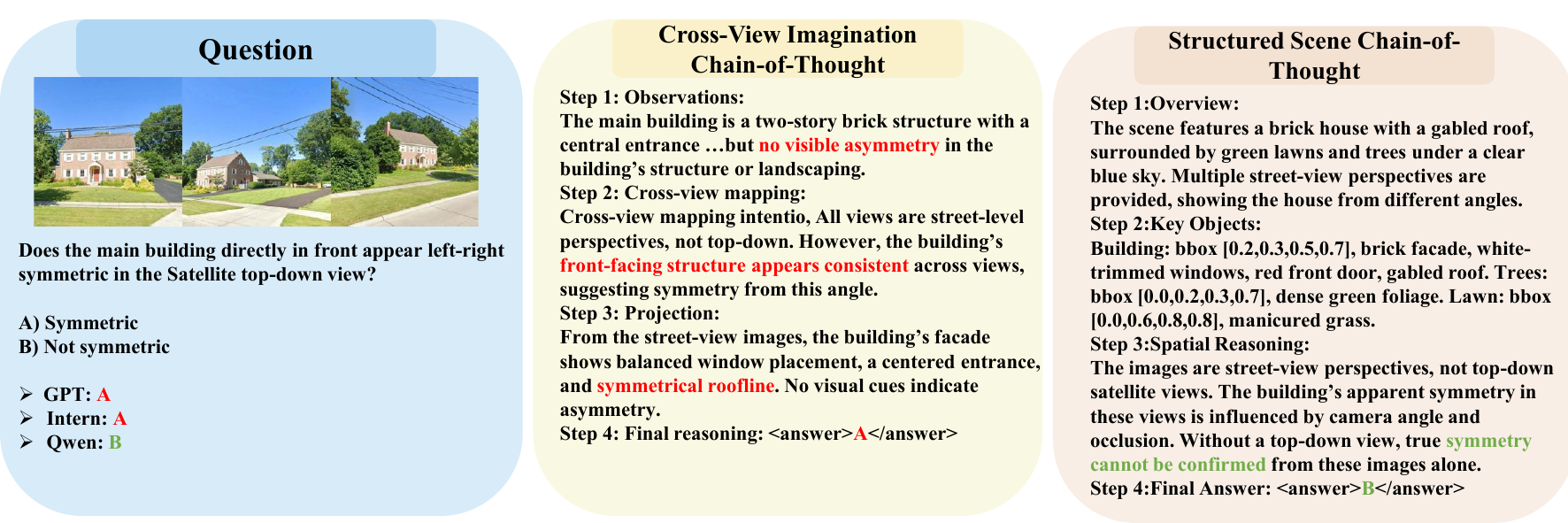}\\[4pt]
\includegraphics[page=6,width=\textwidth]{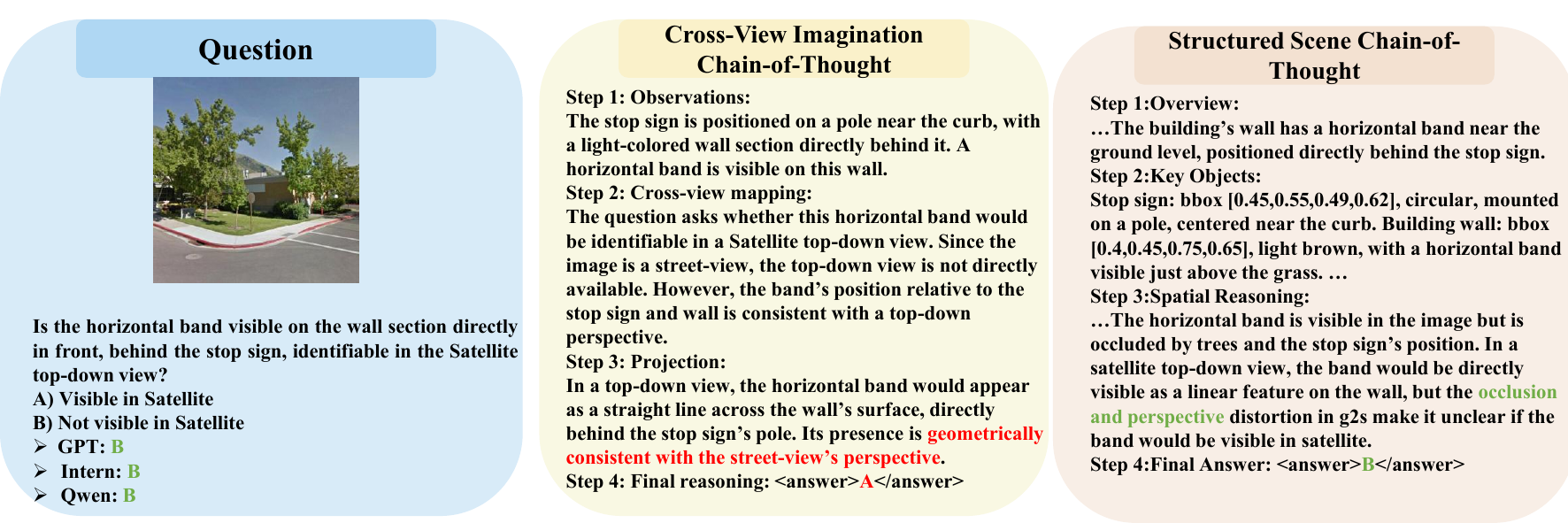}
\caption{
Additional qualitative comparisons between Imagination-CoT
and Structured-CoT on representative CVSBench examples
}
\label{fig:additional_qualitative_results_2}
\end{figure}

\begin{figure}[!htbp]
\centering
\includegraphics[page=7,width=\textwidth]{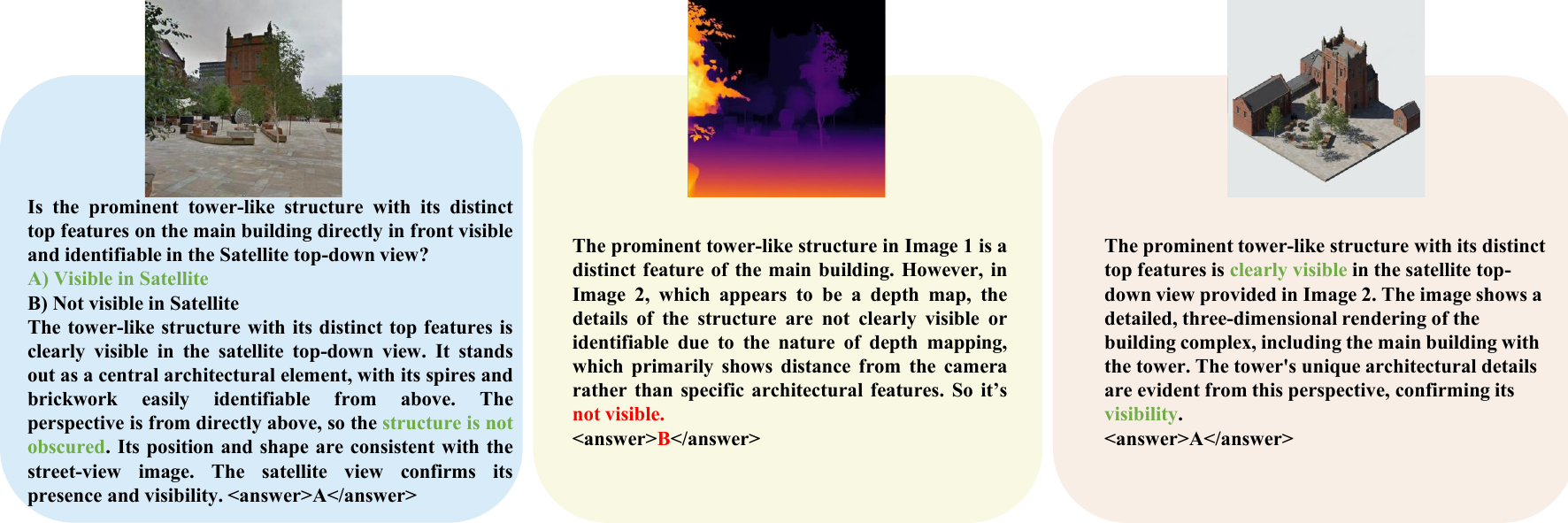}\\[4pt]
\includegraphics[page=8,width=\textwidth]{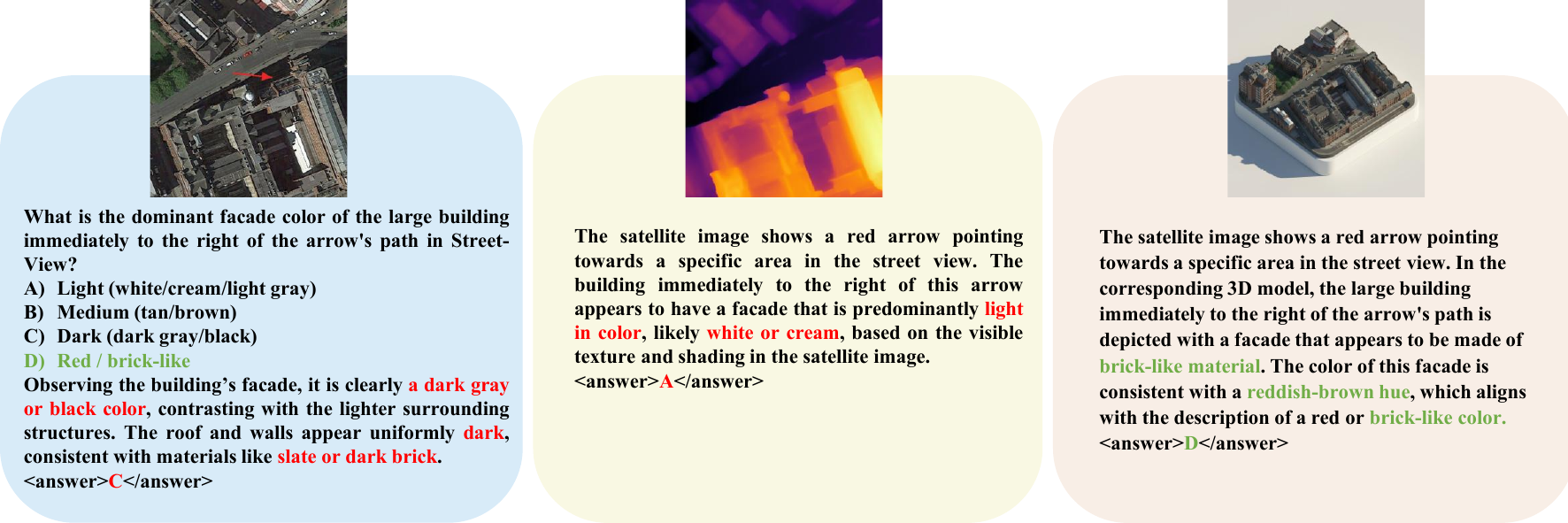}\\[4pt]
\includegraphics[page=9,width=\textwidth]{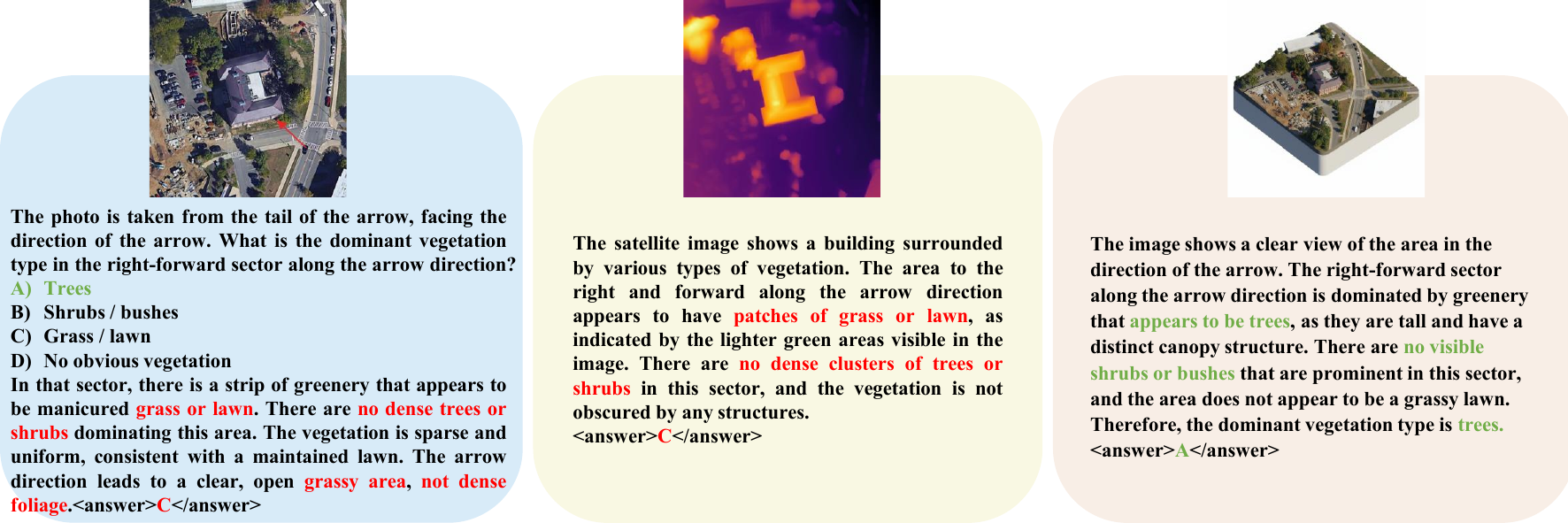}
\caption{
Additional qualitative comparisons between Imagination-CoT
and Structured-CoT on representative CVSBench examples
}
\label{fig:additional_qualitative_results_3}
\end{figure}

We provide additional qualitative comparisons between the
Cross-View Imagination Chain-of-Thought (Imagination-CoT)
and the Structured Scene Chain-of-Thought (Structured-CoT)
reasoning strategies.
Each example includes the input image, the question,
and the reasoning outputs produced by the two reasoning
styles. These examples illustrate how the imagination-based
reasoning explicitly projects visible cues into the
cross-view spatial layout, while the structured reasoning
focuses on object-centric descriptions and spatial relations.
For readability, we present the first nine pages of the visualization results in three groups (Figs.~\ref{fig:additional_qualitative_results_1}--\ref{fig:additional_qualitative_results_3}).

\subsection{Analysis of IoU Failure Rates}
During the dataset construction and quality control pipeline, we evaluated the validity of the generated spatial bounding boxes. We observed a specific type of failure where the pipeline failed to produce valid pixel coordinates, yielding null outputs instead of the expected numerical bounding box formats. These "null coordinate" instances typically occur when the model encounters extreme cross-view ambiguity, severe occlusion, or struggles with format adherence during the spatial grounding process.

Quantitative analysis indicates that this coordinate generation failure rate is exceptionally low. To guarantee the strict formatting consistency and high spatial reliability of our benchmark, all samples containing such invalid null outputs were rigorously filtered out and completely excluded from the final released dataset.

\section{CVSBench Q\&A}

\noindent\textbf{1. For what purpose was the dataset created? Was there a specific task in mind or a specific gap that needed to be filled?}

CVSBench was created to evaluate whether modern Vision--Language Models
(VLMs) can perform robust cross-view spatial reasoning between street-view
and satellite-view observations. Existing spatial benchmarks mainly focus on
indoor objects, simple environments, or limited viewpoint changes, and
therefore do not adequately test cross-view reasoning under complex real-world
urban scenes. CVSBench is designed to fill this gap by unifying cross-view
VQA, cross-view grounding, and viewpoint localization in a single benchmark
built on satellite--street view pairs.

\vspace{0.5em}
\noindent\textbf{2. What do the instances in the dataset represent?}

The benchmark is constructed from paired satellite and street-view imagery.
At the dataset level, CVSBench contains cross-view image groups together with
task annotations for three task families: cross-view VQA, cross-view
grounding, and viewpoint localization. Depending on the subtask, an instance
may contain a single input image, multiple candidate images, a question with
answer options, a target bounding box, or a viewpoint arrow.

\vspace{0.5em}
\noindent\textbf{3. How many instances are there in total, and how are they organized?}

CVSBench contains 3,297 image groups in total, including 2,155
satellite--panorama pairs curated from CVUSA and 1,142 satellite--street-view
pairs extracted from University-1652. The benchmark is organized into
multiple subtasks, including CVUSA-subset G2S (2,260), CVUSA-subset S2G
(5,131), FOV-subset G2S (4,436), FOV-subset S2G (5,741), cross-view grounding
(18,932), View-Arrow (2,543), and View-Image (1,636), which together sum to
40,679 QA or evaluation instances.

\vspace{0.5em}
\noindent\textbf{4. Does the dataset contain all possible instances, or is it a sample from a larger source? If it is a sample, what is the larger source?}

The dataset is a curated sample rather than an exhaustive collection.
CVSBench is built from two existing cross-view datasets: CVUSA and
University-1652. From these larger sources, the authors select high-quality
satellite--street image pairs and then construct task annotations through a
semi-automatic pipeline with human verification, rather than directly using
all source data without filtering.

\vspace{0.5em}
\noindent\textbf{5. Is there a label or target associated with each instance?}

Yes. Each instance has an explicit supervision target appropriate to its task.
For cross-view VQA, the target is the correct answer option. For grounding,
the target is the corresponding object region represented as a bounding box in
the other view. For viewpoint localization, the target is either the correct
viewpoint arrow or the correct street-view image associated with a satellite
arrow cue.

\vspace{0.5em}
\noindent\textbf{6. How was the data associated with each instance acquired and annotated?}

CVSBench is constructed using a semi-automatic annotation pipeline with human
verification. For cross-view VQA, structured prompt templates are combined
with paired images to generate candidate questions using
\texttt{gemini-2.5-\allowbreak flash}, after which invalid or image-independent questions
are filtered and the remaining questions are manually verified. For
viewpoint localization, human annotators mark directional arrows on satellite
images to indicate camera location and viewing orientation. For grounding,
the CVUSA-subset first uses model-generated candidate boxes followed by manual
review and correction, while the FOV-subset uses manual bounding-box
annotation directly.

\vspace{0.5em}
\noindent\textbf{7. What quality-control or verification procedures were used?}

Quality control is centered on human verification. Questions that can be
answered without visual input are filtered out before manual review. The
primary review criteria include cross-view identifiability, semantic
consistency across views, and correction of incorrect answers. Eight professional annotators spend approximately 100 hours on
annotation and verification, and around 30\% of the QA pairs are
manually corrected during this process.

\vspace{0.5em}
\noindent\textbf{8. Does the dataset relate to people or contain potentially sensitive information?}

The benchmark is built from real-world satellite and street-view imagery, so
people, vehicles, and other transient objects may appear incidentally in the
raw images. However, the benchmark design focuses on stable spatial structures
such as buildings, roads, vegetation, viewpoints, and cross-view object
correspondence. In addition, the question-generation constraints explicitly
discourage the use of unstable or temporary objects as anchors, reducing the
reliance on potentially sensitive or non-stationary content.

\vspace{0.5em}
\noindent\textbf{9. How will the dataset be distributed, and what parts of the data will be released?}

According to the paper, the data and code will be released. At the same time,
the appendix specifies that the benchmark is constructed from publicly
available source datasets, namely CVUSA and University-1652, and that the raw
images will not be redistributed. Instead, the release will include the
annotations, viewpoint arrows, cross-view bounding boxes, and question--answer
pairs required to reconstruct the benchmark. Users are expected to obtain the
original images from the official sources and combine them with the released
annotations to build the full benchmark.

\end{document}